\documentclass[11pt]{article}

\usepackage[preprint]{acl}

\usepackage{times}
\usepackage{latexsym}
\usepackage[T1]{fontenc}
\usepackage[utf8]{inputenc}
\usepackage{microtype}
\usepackage{inconsolata}
\usepackage{graphicx}
\usepackage{booktabs}
\usepackage{amsmath,amssymb,amsfonts,amsthm}
\usepackage{bm}
\usepackage{bbm}
\usepackage{multirow}
\usepackage{xcolor}
\usepackage{colortbl}
\usepackage{array}
\usepackage{tabularx}
\usepackage{enumitem}
\usepackage{listings}
\usepackage[ruled,vlined]{algorithm2e}
\SetKwInput{KwInput}{Input}
\SetKwInput{KwOutput}{Output}

\newtheorem{proposition}{Proposition}

\newtheorem{corollary}{Corollary}

\definecolor{oursbg}{RGB}{253,238,240}
\definecolor{simred}{RGB}{165,30,45}
\definecolor{codebg}{RGB}{248,249,250}

\newcommand{\method}{\textsc{Weir}}
\newcommand{\E}{\mathbb{E}}
\newcommand{\Var}{\mathrm{Var}}
\newcommand{\Cov}{\mathrm{Cov}}

\newcommand{\ind}{\mathbb{I}}

\newcommand{\xmark}{$\times$}

\title{\method{}: Watershed Exploration and Drift-Aware Prefix Shrinkage \\ for Multi-Turn Agent Reinforcement Learning}

\author{
  \textbf{Pixel Nomand\textsuperscript{1}},
  \textbf{Elena Voss\textsuperscript{1}},
  \textbf{Marcus Hale\textsuperscript{2}},
  \textbf{Sofia Reyes\textsuperscript{1}}
  \\
  \\
  \textsuperscript{1}University of Wisconsin--Madison \\
  \textsuperscript{2}University of Washington \\
  \texttt{nomand@cs.wisc.edu}
}

\begin{document}
\maketitle

\begin{abstract}
Multi-turn language model agents often receive only a single binary reward at the end of a long interaction trajectory, making it difficult to separate decisive decisions from routine steps or estimate accurate intermediate values. Prefix-tree methods improve sample efficiency by merging trajectories that share action histories, yet two structural flaws remain: naive pooling mixes stale off-policy outcomes into critic targets with fixed weights, and uniform or entropy-driven branching wastes continuations on dead-end states whose outcome is already settled. We introduce \method{} (\emph{Watershed Exploration and Intrinsic-Drift Shrinkage}), an actor-critic reinforcement learning framework grounded in a Doob turn-variance conservation identity. \method{} maintains a cross-iteration canonical action-prefix tree with exponentially discounted visit counts and trains a prefix-conditioned critic toward a minimax drift-aware shrinkage target that adapts from single-trajectory returns at rare nodes to pooled empirical means at dense nodes. Using the same pre-update critic, \method{} allocates rollout budgets across tasks via Neyman stratification $B_q \propto \sqrt{\hat{p}_q(1-\hat{p}_q)}$ and directs within-task branches to \emph{watershed forks} where high survival-headroom uncertainty coincides with steep value cliffs. Across four multi-hop QA benchmarks, ALFWorld, and WebShop on \textsc{Qwen2.5-3B/7B-Instruct}, \method{} achieves $32.18\%$ and $37.85\%$ average QA exact match and $84.83\%$ and $87.81\%$ ALFWorld success rates, outperforming CAPO by $1.50$--$3.74$ points while generating $28$--$29\%$ fewer rollout tokens than chain sampling.
\end{abstract}

\section{Introduction}
\label{sec:intro}

Reinforcement learning with verifiable outcomes enables large language models (LLMs) to interleave chain-of-thought reasoning with multi-turn search, navigation, and tool execution~\citep{yao2023react,jin2025searchr1,wang2025ragen}. Unlike single-turn math benchmarks, agentic tasks span dozens of environment interactions where feedback arrives only at episode completion~\citep{shridhar2021alfworld,yao2022webshop}. Under sparse terminal rewards, trajectory-level group methods such as GRPO~\citep{shao2024deepseekmath} broadcast one scalar advantage across every turn, reinforcing lucky detours in winning episodes and penalizing sound early steps in failed ones~\citep{li2026turnppo,zheng2026pgpo}.

Two families of methods seek finer turn-level credit under a fixed rollout budget. Group and tree-search algorithms organize rollouts into shared-prefix trees or anchor-state clusters to construct intra-tree preference pairs or step potentials without a critic~\citep{feng2025gigpo,yang2025treerpo,ji2025treegrpo,zong2026at2po,wang2026rtmc}. Yet under realistic budgets of six to eight rollouts per task, individual branches often receive too few continuations to stabilize advantage estimates, and late-stage updates can collapse~\citep{zheng2026pgpo,zhu2026gagpo}. Conversely, actor-critic methods such as Turn-PPO~\citep{li2026turnppo} and CAPO~\citep{wang2026capo} train a value network for turn-level Generalized Advantage Estimation (GAE)~\citep{schulman2016high}. However, with $\gamma = \lambda_A = 1$, standard GAE regression targets collapse algebraically to the trajectory's single terminal reward, leaving intermediate states supervised by high-variance one-shot outcomes.

\begin{figure*}[t]
    \centering
    \includegraphics[width=0.98\textwidth]{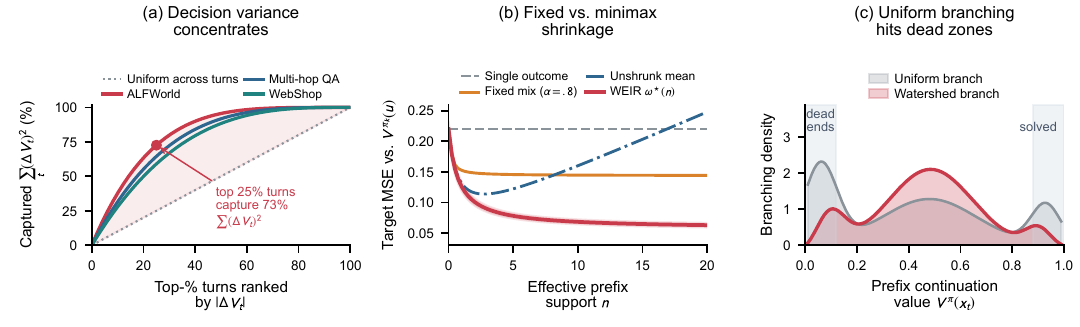}
    \caption{\textbf{Two structural bottlenecks in outcome-based agent RL.} \textbf{(a)} By Doob turn-variance conservation (Proposition~\ref{prop:doob}), outcome uncertainty concentrates in a minority of watershed turns: the top $25\%$ of turns capture $73\%$ of $\sum_t (\Delta V_t)^2$ on ALFWorld. \textbf{(b)} Under policy drift $\kappa{=}0.35$, fixed target mixing ($\alpha{=}0.8$) underuses dense nodes while unshrunk pooling drifts; minimax shrinkage $\omega^\star(n)$ yields uniformly lower target MSE. \textbf{(c)} Uniform branching spends $>40\%$ of continuations on settled dead ends ($V{<}0.1$) or solved prefixes ($V{>}0.9$), whereas watershed branching focuses on undecided forks.}
    \label{fig:motivation}
\end{figure*}

Pooling outcomes across shared action prefixes bridges these paradigms, yet existing designs suffer from two coupled flaws (Figure~\ref{fig:motivation}). First, \emph{target estimation ignores sample precision and policy drift}: fixed blending between current returns and historical prefix statistics either wastes evidence at dense nodes or trusts stale counts from earlier policies at sparse nodes. Second, \emph{rollout allocation ignores where credit is decided}: a Doob martingale conservation law shows that total outcome variance decomposes across turns as the sum of squared value jumps, which concentrates in a few decisive \emph{watershed turns}. Branching uniformly at random~\citep{ji2025treegrpo} or purely by token entropy~\citep{zong2026at2po} frequently expands already-failed dead ends or routine steps whose values barely move.

We present \method{} (\emph{Watershed Exploration and Intrinsic-Drift Shrinkage}), a closed-loop actor-critic framework coupling drift-aware prefix value learning with bilevel watershed allocation. \method{} records rollouts in a persistent canonical action-prefix tree whose visit counts decay exponentially across iterations ($\gamma_{\mathrm{hist}}$). The critic regresses onto a closed-form \emph{minimax shrinkage target} interpolating smoothly from the fresh return at unvisited nodes to the pooled prefix mean at well-supported nodes. Simultaneously, the pre-update critic guides exploration at two levels: across tasks via Neyman stratification $B_q \propto \sqrt{\hat{p}_q(1-\hat{p}_q)}$, and within tasks at watershed forks combining survival-headroom uncertainty $\hat{v}_t(1-\hat{v}_t)$ with steep value transitions $|\bar{V}_k(x_{t+1})-\bar{V}_k(x_t)|$. Our contributions are:
\begin{itemize}[leftmargin=*,itemsep=1.5pt,topsep=2pt]
\item We prove a Doob turn-variance conservation identity for non-Markovian POMDPs, showing why outcome variance concentrates in decisive watershed turns (\S\ref{sec:theory}, Figure~\ref{fig:motivation}a).
\item We propose \method{}, unifying a discounted canonical prefix tree, minimax drift shrinkage, Neyman task budgeting, and watershed branching in one closed loop (\S\ref{sec:method}, Figure~\ref{fig:pipeline}).
\item We derive optimality guarantees for minimax shrinkage, watershed sampling efficiency ($1+\mathrm{CV}^2$), and Neyman stratification, verified numerically and empirically (\S\ref{sec:theory}, \S\ref{sec:analysis}).
\item Across Multi-hop QA, ALFWorld, and WebShop at 3B and 7B scales, \method{} outperforms CAPO by $1.50$--$3.74$ points using $28$--$29\%$ fewer tokens than chain sampling (\S\ref{sec:experiments}).
\end{itemize}

\section{Related Work}
\label{sec:related}

\paragraph{Multi-turn credit assignment under sparse rewards.}
LLM RL uses hierarchical or token-level critics~\citep{zhou2024archer,yuan2025vcppo,yue2025vapo}, while critic-free methods extend REINFORCE and GRPO to search and embodied tasks~\citep{ahmadian2024back,shao2024deepseekmath,yu2025dapo,jin2025searchr1,sun2025zerosearch,wei2025reinforcing}. To replace uniform trajectory credit, recent 2025--2026 methods derive step-level signals via anchor-state grouping~\citep{feng2025gigpo}, potential differences~\citep{zheng2026pgpo}, grouped TD proxies~\citep{zhu2026gagpo}, influence graphs~\citep{ren2026iapo}, transition rubrics~\citep{zhang2026trca}, role typing~\citep{xu2026triage}, checkpoint TD residuals~\citep{ma2026factor}, or lightweight shared critics~\citep{zhou2025sweetrl,chen2026vetta}. In actor-critic RL, Turn-PPO~\citep{li2026turnppo} and CAPO~\citep{wang2026capo} align updates with interaction turns, while SPPO~\citep{wang2026sppo} uses a sequence-level scalar critic; each trajectory supplies only one terminal target. \method{} pools outcomes across trees and prior updates via drift-discounted shrinkage.

\paragraph{Tree-structured rollouts and adaptive allocation.}
Tree methods reuse shared prefixes for rollout expansion and credit assignment~\citep{kazemnejad2025vineppo,yang2025treerpo,tran2025tempo,ji2025treegrpo,wang2026rtmc}. TreeOPSD~\citep{zou2026treeopsd} backs terminal rewards through shared-prefix trees into bounded posterior distillation targets. At the prompt level, GRESO~\citep{zheng2025greso}, PCL~\citep{gao2025pcl}, ARCUS~\citep{okonkwo2026arcus}, and CoBA-RL~\citep{yao2026cobarl} prioritize intermediate difficulty, while ARPO~\citep{dong2025arpo}, AT$^2$PO~\citep{zong2026at2po}, and ATPO~\citep{cao2026atpo} expand high-entropy or high-Bellman-error states, IGRPO~\citep{zhang2026igrpo} scores branches by answer information gain, ProGPO~\citep{fan2026progpo} tracks progress reliability, and TRACE~\citep{zou2026trace} allocates budgets with an auxiliary predictor for group-relative updates. SARA~\citep{nomand2026sara} uses sequential stopping, while PAIR~\citep{nomand2026pair} corrects pair inclusion under adaptive allocation. \method{} instead unifies turn-level value baselines and bilevel allocation inside one prefix critic, replacing 3-bin quotas and entropy heuristics with Neyman stratification and watershed cliff selection.

\section{Preliminaries}
\label{sec:prelim}

\paragraph{Multi-turn agentic POMDP.}
For task $q \sim \mathcal{D}$ with initial observation $o_0$, an agent generates a trajectory $\tau = (q, o_0, (c_t, a_t, o_t)_{t=1}^{T_\tau})$, where $c_t$ is reasoning, $a_t$ is a parsed action, and $o_t$ is the observation. At turn $t$, policy $\pi_\theta$ emits $(c_t, a_t)$ given history $h_t = (q, o_0, (c_i, a_i, o_i)_{i<t})$. The environment yields only a terminal reward $R(\tau) \in [0, 1]$, with $r_t = 0$ ($t < T_\tau$) and $r_{T_\tau} = R(\tau)$.

\paragraph{Single-trajectory collapse of undiscounted GAE.}
With $\gamma = \lambda_A = 1$, pre-update critic $\bar{V}$ and $\bar{V}(h_{T_\tau+1}) = 0$, the GAE return target telescopes identically:
\begin{equation}
\begin{aligned}
G_t^\lambda &= \bar{V}(h_t) + \sum_{j=t}^{T_\tau} \big[r_j + \bar{V}(h_{j+1}) - \bar{V}(h_j)\big] \\
&= R(\tau).
\end{aligned}
\label{eq:gae_collapse}
\end{equation}
Every turn along a single trajectory therefore regresses onto the identical noisy scalar $R(\tau)$, causing full-history critics to memorize surface reasoning tokens $c_t$ rather than underlying decision utility.

\section{\method{}}
\label{sec:method}

\begin{figure*}[t]
    \centering
    \includegraphics[width=0.98\textwidth]{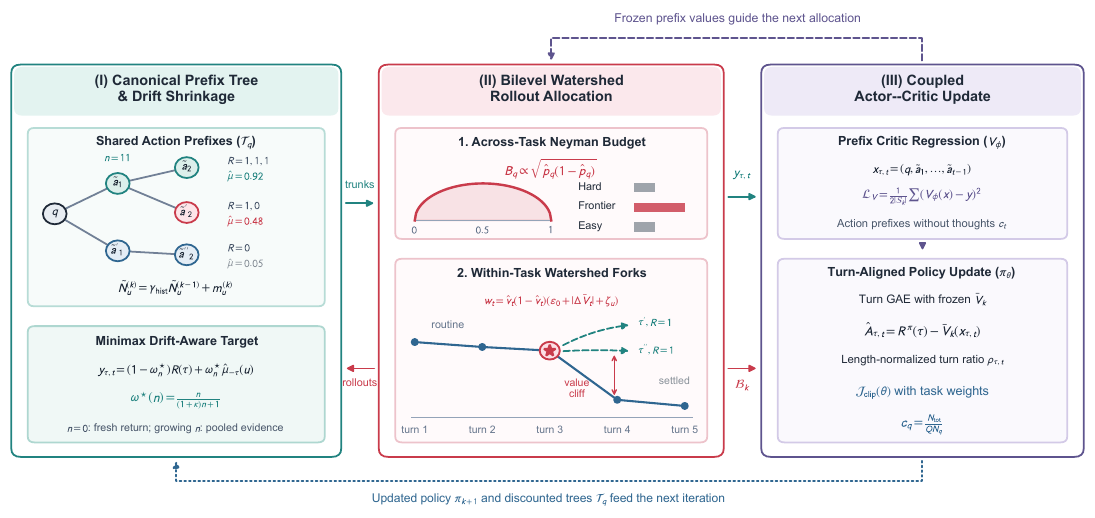}
    \caption{\textbf{Overview of \method{}.} \textbf{(I)} Completed trajectories are canonicalized by ordered actions into a persistent prefix tree $\mathcal{T}_q$ with exponentially discounted counts $\tilde{N}_u^{(k)}$, producing minimax drift-shrinkage targets $y_{\tau,t}$. \textbf{(II)} The pre-update critic $\bar{V}_k$ scores trunk prefixes to assign task budgets via Neyman stratification $B_q \propto \sqrt{\hat{p}_q(1-\hat{p}_q)}$ and selects branch points at watershed forks combining survival uncertainty and value cliffs. \textbf{(III)} The prefix critic $V_\phi$ fits $y_{\tau,t}$ while the actor $\pi_\theta$ optimizes a turn-aligned, task-balanced clipped surrogate using turn-level GAE.}
    \label{fig:pipeline}
\end{figure*}

\method{} couples a canonical prefix value tree with bilevel watershed allocation in a closed loop (Figure~\ref{fig:pipeline}).

\subsection{Canonical Prefix Tree and Drift-Aware Shrinkage}
\label{sec:method-tree}

\paragraph{Action-prefix canonicalization.}
Different reasoning traces $c_t$ often emit identical action sequences. We map each history $h_t$ to its canonical action prefix $x_t = g(h_t) = (q, \tilde{a}_1, \dots, \tilde{a}_{t-1})$, with $x_1 = (q)$. Using normalized exact matching for embodied/web actions and parent-constrained E5-base-v2 semantic merging ($\cos \ge \eta{=}0.95$) for QA queries~\citep{wang2022text}, rollouts populate a persistent Action-Prefix Tree $\mathcal{T}_q$. Conditioning $V_\phi(x_t)$ on $x_t$ aligns its predictions with pooled outcomes and removes variance from verbose thoughts $c_t$.

\paragraph{Discounted cross-iteration statistics.}
Each node $u \in \mathcal{T}_q$ maintains exponentially discounted outcome counts and return sums across iterations $k$:
\begin{equation}
\begin{aligned}
\tilde{N}_u^{(k)} &= \gamma_{\mathrm{hist}} \tilde{N}_u^{(k-1)} + m_u^{(k)}, \\
\tilde{S}_u^{(k)} &= \gamma_{\mathrm{hist}} \tilde{S}_u^{(k-1)} + \!\sum_{\tau \in \mathcal{B}_{k,q}} \ind[u \in W(\tau)] R(\tau),
\end{aligned}
\label{eq:disc_stats}
\end{equation}
where $W(\tau)$ is the node path of $\tau$, $m_u^{(k)} = \sum_{\tau \in \mathcal{B}_{k,q}} \ind[u \in W(\tau)]$, and $\gamma_{\mathrm{hist}} \in (0,1]$ down-weights older policies. Excluding $\tau$ itself prevents self-leakage:
\begin{equation}
\begin{aligned}
n_{-\tau}(u) &= \tilde{N}_u^{(k)} - \ind[u \in W(\tau)], \\
\hat{\mu}_{-\tau}(u) &= \frac{\tilde{S}_u^{(k)} - \ind[u \in W(\tau)] R(\tau)}{n_{-\tau}(u)},
\end{aligned}
\label{eq:loo_prefix}
\end{equation}
whenever $n_{-\tau}(u) > 0$.

\paragraph{Minimax drift-aware precision shrinkage.}
Modeling $\hat{\mu}_{-\tau}(u)$ with variance $\sigma_u^2 / n$ and bounded policy-shift bias $|b_u| \le \sqrt{\kappa}\,\sigma_u$ relative to $V^{\pi_k}(u)$, minimizing worst-case MSE gives the unique shrinkage weight (Proposition~\ref{prop:shrinkage}):
\begin{equation}
\begin{aligned}
\omega^\star(n) &= \frac{n}{(1+\kappa)n + 1}, \\
y_{\tau,t} &= \big(1 - \omega^\star_{n}\big) R(\tau) + \omega^\star_{n}\,\hat{\mu}_{-\tau}(u_{\tau,t}),
\end{aligned}
\label{eq:shrinkage}
\end{equation}
where $\omega^\star_n := \omega^\star(n_{-\tau}(u_{\tau,t}))$ and $\omega^\star(0) = 0$. The critic minimizes $\mathcal{L}_V(\phi) = \frac{1}{2|\mathcal{S}_k|}\sum_{(x,y)\in \mathcal{S}_k} (V_\phi(x) - y)^2$ over frozen targets (plus terminal answer samples in QA).

\subsection{Bilevel Watershed Rollout Allocation}
\label{sec:method-alloc}

Under frozen policy $\pi_k$ and pre-update critic $\bar{V}_k$, \method{} samples $M_0$ trunk trajectories per task and distributes $B_{\mathrm{tot}} = Q \bar{b}$ expansion branches across two stages.

\paragraph{Level 1: Across-task Neyman stratification.}
Task $q$'s success rate is estimated by averaging clipped trunk-prefix values $\hat{p}_q = \frac{1}{|P_q|}\sum_{x \in P_q}\mathrm{clip}(\bar{V}_k(x), \epsilon_p, 1{-}\epsilon_p)$, with uncertainty $s_q = \sqrt{\hat{p}_q(1-\hat{p}_q)}$. To minimize batch gradient variance under task balancing (Proposition~\ref{prop:neyman}), \method{} allocates budgets proportional to $s_q$ within $[b_{\min}, b_{\max}]$:
\begin{equation}
\begin{aligned}
B_q = \mathrm{round}\!\Big(\mathrm{clip}\!\big(&b_{\min} + \tilde{B}_{\mathrm{ex}}\frac{s_q}{\sum_{q'} s_{q'}}, \\
&b_{\min},\, b_{\max}\big)\Big),
\end{aligned}
\label{eq:neyman}
\end{equation}
where $\tilde{B}_{\mathrm{ex}} = B_{\mathrm{tot}} - Q b_{\min}$, adjusting integer rounding to preserve $\sum_q B_q = Q \bar{b}$ ($\bar{b}{=}4, b_{\min}{=}2, b_{\max}{=}7$).

\paragraph{Level 2: Within-task watershed turn selection.}
By Proposition~\ref{prop:doob}, decision credit concentrates where continuation values shift sharply while outcome uncertainty remains unresolved. Each non-terminal turn $(\tau, t)$ receives the \emph{watershed score}:
\begin{equation}
\begin{aligned}
w_{\tau,t} = \hat{v}_{\tau,t}(1{-}\hat{v}_{\tau,t})\Big(&\varepsilon_0 + |\bar{V}_k(x_{\tau,t+1}) - \bar{V}_k(x_{\tau,t})| \\
&+ \frac{\lambda_\zeta}{\sqrt{n(\tau,t){+}1}}\Big),
\end{aligned}
\label{eq:watershed}
\end{equation}
where $\hat{v}_{\tau,t} = \mathrm{clip}(\bar{V}_k(x_{\tau,t}), 0, 1)$, $\varepsilon_0{=}0.05$, and $\lambda_\zeta{=}0.15$. We sample $B_q$ branch origins proportionally to $w_{\tau,t}$ without replacement, restore their environment states and histories, and roll out continuations in parallel under $\pi_k$.

\subsection{Coupled Turn-Aligned Policy Optimization}
\label{sec:method-actor}

Using denoised prefix baselines $\bar{V}_k(x_{\tau,t})$ and on-policy rewards $r^\pi_{\tau,t}$ (with format gate $F(\tau)R(\tau)$ in QA), turn-level GAE with $\gamma{=}\lambda_A{=}1$ and $\bar{V}_k(x_{\tau,T_\tau+1}){=}0$ is
\begin{equation}
\begin{aligned}
\delta_{\tau,t} &= r^\pi_{\tau,t} + \gamma \bar{V}_k(x_{\tau,t+1}) - \bar{V}_k(x_{\tau,t}), \\
\hat{A}_{\tau,t} &= \sum_{\ell=0}^{T_\tau - t} (\gamma \lambda_A)^\ell \delta_{\tau,t+\ell}.
\end{aligned}
\label{eq:turn_gae}
\end{equation}
For turn $t$'s generated tokens $z_{\tau,t}$, the length-normalized geometric importance ratio~\citep{zheng2025gspo,li2026turnppo} is
\begin{equation}
\begin{aligned}
\rho_{\tau,t}(\theta) &= \exp\!\Big(\frac{1}{|z_{\tau,t}|}\sum_{j=1}^{|z_{\tau,t}|}\log r_{\tau,t,j}(\theta)\Big), \\
r_{\tau,t,j}(\theta) &= \frac{\pi_\theta(z_{\tau,t,j} \mid h_{\tau,t}, z_{\tau,t,<j})}{\pi_k(z_{\tau,t,j} \mid h_{\tau,t}, z_{\tau,t,<j})}.
\end{aligned}
\label{eq:gs_ratio}
\end{equation}
To keep task contributions unbiased under Neyman counts $N_q = M_0 + B_q$, each trajectory uses weight $c_q = N_{\mathrm{tot}}/(Q N_q)$ in the clipped objective:
\begin{equation}
\begin{aligned}
\mathcal{J}_\pi(\theta) &= \E_{\tau \sim \mathcal{B}_k}\!\Big[\frac{c_{q(\tau)}}{|I_\tau|} \sum_{j \in I_\tau} \min\!\big(\tilde{\rho}_{\tau,j}\hat{A}_{\tau,t(j)}, \\
&\quad \mathrm{clip}(\tilde{\rho}_{\tau,j}, 1{-}\epsilon_l, 1{+}\epsilon_h)\hat{A}_{\tau,t(j)}\big)\Big] \\
&\quad - \beta_{\mathrm{KL}} \mathcal{L}_{\mathrm{KL}}(\theta),
\end{aligned}
\label{eq:actor_obj}
\end{equation}
where $\tilde{\rho}_{\tau,j}$ retains token-level gradients via stop-gradient scaling~\citep{zheng2025gspo} and $I_\tau$ masks observation tokens (Algorithm~\ref{alg:weir}).

\begin{algorithm}[t]
\footnotesize
\DontPrintSemicolon
\SetAlgoLined
\caption{\method{} online training iteration $k$}
\label{alg:weir}
\KwInput{Batch $\mathcal{Q}_k$ ($|\mathcal{Q}_k|{=}Q$), sampling policy $\pi_k$, pre-update critic $\bar{V}_k$, discounted trees $\{\mathcal{T}_q\}$, budget $\bar{b}$}
Sample $M_0$ trunk trajectories per task $q \in \mathcal{Q}_k$ under frozen $\pi_k$\;
Evaluate trunk prefixes with $\bar{V}_k$; assign Neyman budgets $B_q$ via \eqref{eq:neyman}\;
Score trunk turns via \eqref{eq:watershed}; sample $B_q$ branch origins\;
Restore environment states at selected origins and sample continuations under $\pi_k$\;
Insert trajectories into $\{\mathcal{T}_q\}$; update $(\tilde{N}_u^{(k)}, \tilde{S}_u^{(k)})$ via \eqref{eq:disc_stats}\;
Compute leave-one-out shrinkage targets $y_{\tau,t}$ via \eqref{eq:loo_prefix}--\eqref{eq:shrinkage}\;
Compute turn GAE $\hat{A}_{\tau,t}$ via \eqref{eq:turn_gae} and task weights $c_q = N_{\mathrm{tot}}/(Q N_q)$\;
Update critic $V_\phi$ on $\mathcal{L}_V(\phi)$ and actor $\pi_\theta$ on $\mathcal{J}_\pi(\theta)$ in \eqref{eq:actor_obj}\;
\end{algorithm}

\section{Theoretical Analysis}
\label{sec:theory}

Full proofs and exact numerical verification in a non-Markovian POMDP appear in Appendix~\ref{app:proofs}.

\begin{proposition}[Doob Turn-Variance Conservation]
\label{prop:doob}
For binary rewards $R(\tau)\in\{0,1\}$, let $M_t = V^\pi(x_t) = \E_\pi[R(\tau) \mid x_t]$ for $1 \le t \le T$ and $M_{T+1} = R(\tau)$, where $x_t = (q, \tilde{a}_1, \dots, \tilde{a}_{t-1})$ is the canonical action prefix in an arbitrary non-Markovian POMDP. Then $\{M_t\}_{t=1}^{T+1}$ is a Doob martingale with respect to the canonical action filtration, and
\begin{equation}
V^\pi(q)\big(1 - V^\pi(q)\big) = \sum_{t=1}^{T} \E_\pi\!\left[\big(M_{t+1} - M_t\big)^2\right].
\label{eq:doob}
\end{equation}
\end{proposition}

Proposition~\ref{prop:doob} requires no Markov assumption: latent thoughts $c_t$ and observations $o_t$ are marginalized, yet total outcome uncertainty is strictly conserved across action jumps.

\begin{proposition}[Minimax Drift-Aware Shrinkage]
\label{prop:shrinkage}
Let $R(\tau)$ be unbiased with variance $\sigma_u^2$ and $\hat{\mu}_{-\tau}(u)$ be an independent pooled estimate with count $n > 0$, variance $\sigma_u^2/n$, and drift bias $|b_u| \le \sqrt{\kappa}\,\sigma_u$. Among linear combinations $y_\omega = (1-\omega)R(\tau) + \omega \hat{\mu}_{-\tau}(u)$, the unique minimax-MSE weight is $\omega^\star(n) = \frac{n}{(1+\kappa)n + 1}$, yielding
\begin{equation}
\begin{aligned}
\sup_{|b_u|\le \sqrt{\kappa}\sigma_u} \E\big[(y_{\omega^\star} {-} V^{\pi_k}(u))^2\big]
&= \sigma_u^2\,\frac{\kappa n + 1}{(1{+}\kappa)n + 1} \\
&< \sigma_u^2.
\end{aligned}
\end{equation}
\end{proposition}

\begin{proposition}[Watershed Branching Efficiency]
\label{prop:watershed}
Let $\nu_t \ge 0$ be the expected variance reduction from branching at turn $t$ along a trunk, and let $w_t \propto \nu_t$. Relative to uniform turn sampling ($\E_{\mathrm{unif}}[\nu_t]$), watershed sampling achieves
\begin{equation}
\frac{\E_{w}[\nu_t]}{\E_{\mathrm{unif}}[\nu_t]} = 1 + \mathrm{CV}^2(\nu),
\end{equation}
where $\mathrm{CV}(\nu)$ is the coefficient of variation of $\nu_t$ across eligible turns.
\end{proposition}

\begin{proposition}[Neyman Rollout Stratification]
\label{prop:neyman}
If task $q \in \{1,\dots,Q\}$ has binary success rate $p_q$ and trajectory gradient variance $\Var(g_{q,i}) = c\,p_q(1-p_q)$, minimizing $\Var(\hat{g})$ of the task-balanced estimator $\hat{g} = \frac{1}{Q}\sum_{q=1}^Q \frac{1}{N_q}\sum_{i=1}^{N_q} g_{q,i}$ subject to $\sum_q N_q = N_{\mathrm{tot}}$ yields the unique optimum $N_q^\star \propto \sqrt{p_q(1-p_q)}$, strictly dominating uniform and 3-bin allocations whenever $p_q(1-p_q)$ varies across tasks.
\end{proposition}

\section{Experiments}
\label{sec:experiments}

\subsection{Experimental Setup}
\label{sec:setup}

\paragraph{Benchmarks and backbones.}
We train \textsc{Qwen2.5-3B-Instruct} and \textsc{Qwen2.5-7B-Instruct}~\citep{yang2024qwen25} actors (with a 3B critic at both scales; scaling to 1.5B, 14B, and \textsc{Qwen3-4B}~\citep{yang2025qwen3} is in Appendix~\ref{app:extra}). \textbf{(1) Multi-hop QA:} Following~\citet{jin2025searchr1} and~\citet{ji2025treegrpo}, we train on HotpotQA~\citep{yang2018hotpotqa} with an E5-base-v2 Wikipedia retriever~\citep{wang2022text} (3 passages/query, up to 4 searches plus answer) and evaluate exact match (EM) on held-out HotpotQA alongside transfer to 2WikiMultiHopQA~\citep{ho20202wikimultihopqa}, MuSiQue~\citep{trivedi2022musique}, and Bamboogle~\citep{press2023measuring}. \textbf{(2) ALFWorld:} 1,000 train and 134 test tasks up to 40 turns~\citep{shridhar2021alfworld}. \textbf{(3) WebShop:} 1,024 train and 256 test tasks up to 15 turns~\citep{yao2022webshop}. ALFWorld and WebShop average three random seeds.

\paragraph{Baselines and budgets.}
We compare against critic-free group/tree methods: GRPO~\citep{shao2024deepseekmath}, GiGPO~\citep{feng2025gigpo}, Tree-GRPO~\citep{ji2025treegrpo}, AT$^2$PO~\citep{zong2026at2po}, and PGPO~\citep{zheng2026pgpo}; and actor-critic methods: PPO~\citep{schulman2017proximal}, SPPO~\citep{wang2026sppo}, Turn-PPO~\citep{li2026turnppo}, and CAPO~\citep{wang2026capo}. Published Search-R1~\citep{jin2025searchr1} and StepSearch~\citep{zheng2025stepsearch} results serve as external references due to differing data or step supervision. Multi-rollout methods use 6 trajectories/task on QA and 8 on ALFWorld/WebShop. \method{} uses $M_0{=}2/4$ trunks and $\bar{b}{=}4$ average branches (matching 6/8 trajectories while generating $0.71$--$0.72\times$ the tokens of independent chains), with $\gamma_{\mathrm{hist}}{=}0.95$, $\kappa{=}0.4$, $\lambda_\zeta{=}0.15$, and $\eta{=}0.95$ (25/30 critic warm-up steps included inside the 150/120 update budget; Appendix~\ref{app:setup}).

\subsection{Main Results}
\label{sec:main}

\begin{table*}[t]
\centering
\footnotesize
\setlength{\tabcolsep}{4.8pt}
\renewcommand{\arraystretch}{0.95}
\caption{\textbf{Main comparison on Multi-hop QA (EM, \%), ALFWorld (Success, \%), and WebShop (Success, \%).} \textbf{QA Avg.} averages the four QA benchmarks; \textbf{Tok.$\times$} denotes generated rollout tokens relative to independent-chain GRPO at matched trajectory counts (6 for QA, 8 for ALFWorld/WebShop). $\dagger$: external references with different training data/supervision. Best comparable result per column in \textbf{bold}.}
\label{tab:main}
\resizebox{\textwidth}{!}{\begin{tabular}{l cccc c cc c}
\toprule
 & \multicolumn{5}{c}{\textbf{Multi-hop QA (EM, \% $\uparrow$)}} & \textbf{ALFWorld} & \textbf{WebShop} & \\
\cmidrule(lr){2-6}
\textbf{Method} & \textbf{HotpotQA} & \textbf{2Wiki} & \textbf{MuSiQue} & \textbf{Bamboogle} & \textbf{QA Avg.} & \textbf{Succ.\ (\%)} & \textbf{Succ.\ (\%)} & \textbf{Tok.$\times$} \\
\midrule
\multicolumn{9}{c}{\textit{\textbf{Qwen2.5-3B-Instruct}}} \\
\midrule
Search-R1~\citep{jin2025searchr1}$^\dagger$ & 30.40 & 29.30 & 12.00 & 24.00 & 23.93 & --- & --- & --- \\
StepSearch~\citep{zheng2025stepsearch}$^\dagger$ & 34.50 & 32.00 & 17.40 & 34.40 & 29.58 & --- & --- & --- \\
\midrule
GRPO~\citep{shao2024deepseekmath} & 30.80 & 30.30 & 9.65 & 24.80 & 23.89 & 75.87 & 66.87 & 1.00$\times$ \\
GiGPO~\citep{feng2025gigpo} & 33.10 & 32.40 & 11.40 & 28.80 & 26.43 & 78.61 & 72.40 & 1.00$\times$ \\
Tree-GRPO~\citep{ji2025treegrpo} & 30.20 & 29.10 & 8.95 & 29.10 & 24.34 & 61.94 & 60.20 & 0.74$\times$ \\
AT$^2$PO~\citep{zong2026at2po} & 35.40 & 34.10 & 12.85 & 30.40 & 28.19 & 79.85 & 74.13 & 0.76$\times$ \\
PGPO~\citep{zheng2026pgpo} & 35.80 & 34.60 & 13.10 & 31.20 & 28.68 & 80.60 & 74.93 & 1.00$\times$ \\
\midrule
PPO~\citep{schulman2017proximal} & 31.20 & 31.30 & 9.75 & 26.40 & 24.66 & 30.85 & 64.53 & 0.17$\times$ \\
SPPO~\citep{wang2026sppo} & 33.90 & 32.90 & 11.60 & 27.20 & 26.40 & 71.64 & 73.10 & 1.00$\times$ \\
Turn-PPO~\citep{li2026turnppo} & 34.60 & 33.70 & 12.20 & 28.00 & 27.13 & 74.38 & 75.80 & 1.00$\times$ \\
CAPO~\citep{wang2026capo} & 36.20 & 35.10 & 13.45 & 31.20 & 28.99 & 81.09 & 76.27 & 1.00$\times$ \\
\midrule
\rowcolor{oursbg} \textbf{\method{} (ours)} & \textbf{39.80} & \textbf{38.40} & \textbf{15.30} & \textbf{35.20} & \textbf{32.18} & \textbf{84.83} & \textbf{78.47} & 0.71$\times$ \\
\midrule
\multicolumn{9}{c}{\textit{\textbf{Qwen2.5-7B-Instruct}}} \\
\midrule
Search-R1~\citep{jin2025searchr1}$^\dagger$ & 39.40 & 31.20 & 18.10 & 38.40 & 31.78 & --- & --- & --- \\
StepSearch~\citep{zheng2025stepsearch}$^\dagger$ & 38.60 & 36.60 & 22.60 & 40.00 & 34.45 & --- & --- & --- \\
\midrule
GRPO~\citep{shao2024deepseekmath} & 41.00 & 39.00 & 16.70 & 38.40 & 33.78 & 79.10 & 78.57 & 1.00$\times$ \\
GiGPO~\citep{feng2025gigpo} & 41.60 & 39.20 & 17.30 & 40.80 & 34.73 & 82.34 & 79.20 & 1.00$\times$ \\
Tree-GRPO~\citep{ji2025treegrpo} & 34.25 & 33.90 & 13.45 & 33.60 & 28.80 & 66.67 & 70.73 & 0.75$\times$ \\
AT$^2$PO~\citep{zong2026at2po} & 41.90 & 39.40 & 17.85 & 41.60 & 35.19 & 83.08 & 79.67 & 0.77$\times$ \\
PGPO~\citep{zheng2026pgpo} & 42.10 & 39.50 & 18.05 & 42.40 & 35.51 & 83.83 & 80.13 & 1.00$\times$ \\
\midrule
PPO~\citep{schulman2017proximal} & 40.40 & 37.50 & 17.50 & 39.20 & 33.65 & 56.22 & 69.50 & 0.17$\times$ \\
SPPO~\citep{wang2026sppo} & 40.80 & 37.90 & 16.90 & 40.80 & 34.10 & 78.36 & 78.90 & 1.00$\times$ \\
Turn-PPO~\citep{li2026turnppo} & 39.00 & 36.20 & 13.50 & 42.00 & 32.68 & 80.10 & 79.73 & 1.00$\times$ \\
CAPO~\citep{wang2026capo} & 42.30 & 39.60 & 18.20 & 42.40 & 35.63 & 84.33 & 80.40 & 1.00$\times$ \\
\midrule
\rowcolor{oursbg} \textbf{\method{} (ours)} & \textbf{44.30} & \textbf{41.20} & \textbf{19.90} & \textbf{46.00} & \textbf{37.85} & \textbf{87.81} & \textbf{81.90} & 0.72$\times$ \\
\bottomrule
\end{tabular}
}
\end{table*}

\paragraph{Multi-hop QA.}
Table~\ref{tab:main} shows \method{} leads every QA benchmark at both scales among matched baselines: $32.18\%$ average EM at 3B ($+3.19$ over CAPO, $+5.05$ over Turn-PPO, $+8.29$ over GRPO) and $37.85\%$ at 7B ($+2.22$ over CAPO, $+5.17$ over Turn-PPO, $+4.07$ over GRPO), while reducing rollout tokens by $28$--$29\%$ ($0.71$--$0.72\times$). Gains persist across in-domain HotpotQA ($44.30\%$ at 7B) and compositional transfer sets (MuSiQue $19.90\%$, Bamboogle $46.00\%$). Against StepSearch, which uses MuSiQue process supervision, \method{} leads by $+5.70$ points on HotpotQA and $+4.60$ on 2Wiki at 7B.

\paragraph{Embodied and web tasks.}
On ALFWorld and WebShop, \method{} reaches $84.83\%$ / $78.47\%$ at 3B and $87.81\%$ / $81.90\%$ at 7B, exceeding CAPO by $+3.74$ / $+2.20$ points (3B) and $+3.48$ / $+1.50$ points (7B). Category-level breakdowns (Appendix~\ref{app:extra}, Table~\ref{tab:env_detail}) show the largest gains on long-horizon Pick2 ($+4.3$ over CAPO) and Look ($+3.7$), while reducing average ALFWorld turns to $13.7$ (vs.\ $16.9$ for Turn-PPO and $17.8$ for GRPO).

\subsection{Ablation Studies}
\label{sec:ablation}

Table~\ref{tab:ablation} isolates each component across both scales. Replacing minimax shrinkage $\omega^\star(n)$ with fixed mixing ($\alpha{=}0.80$) drops 7B QA by $1.03$ and ALFWorld by $1.49$; disabling historical discounting ($\gamma_{\mathrm{hist}}{=}1.0, \kappa{=}0$) costs $0.81$ and $1.24$; removing cross-step retention loses $1.65$ and $2.69$; and using a full-history single-outcome critic loses $3.75$ and $5.97$. Replacing watershed branching with uniform branching drops 7B QA by $0.97$ and ALFWorld by $1.62$, while entropy-guided branching trails by $0.80$ and $1.37$. Removing the survival--headroom gate $\hat{v}_t(1-\hat{v}_t)$ drops ALFWorld by $2.19$. Finally, 3-bin tertile budgeting loses $0.45$ QA and $0.70$ ALFWorld, and dropping task weights $c_q$ costs $0.57$ and $0.87$.

\begin{table}[t]
\centering
\footnotesize
\setlength{\tabcolsep}{4.0pt}
\renewcommand{\arraystretch}{0.94}
\caption{\textbf{Ablation study} on 3B and 7B actors across Multi-hop QA Avg.\ (EM, \%) and ALFWorld (Success, \%).}
\label{tab:ablation}
\resizebox{\columnwidth}{!}{\begin{tabular}{l cc cc}
\toprule
 & \multicolumn{2}{c}{\textbf{3B Actor}} & \multicolumn{2}{c}{\textbf{7B Actor}} \\
\cmidrule(lr){2-3}\cmidrule(lr){4-5}
\textbf{Configuration} & \textbf{QA} & \textbf{ALF.} & \textbf{QA} & \textbf{ALF.} \\
\midrule
\rowcolor{oursbg} \textbf{\method{} (full)} & \textbf{32.18} & \textbf{84.83} & \textbf{37.85} & \textbf{87.81} \\
\midrule
Fixed mixing ($\alpha{=}0.80$, no count adaptation) & 31.15 & 83.40 & 36.82 & 86.32 \\
No drift discounting ($\gamma_{\mathrm{hist}}{=}1.0$, $\kappa{=}0$) & 31.38 & 83.65 & 37.04 & 86.57 \\
Current-batch tree only (no cross-step retention) & 30.62 & 82.10 & 36.20 & 85.12 \\
Full-history critic (no prefix pooling, Turn-PPO style) & 28.45 & 76.90 & 34.10 & 81.84 \\
\midrule
Uniform within-task branching (keep Neyman task budget) & 31.20 & 83.33 & 36.88 & 86.19 \\
Entropy-guided branching (replace watershed score) & 31.35 & 83.58 & 37.05 & 86.44 \\
w/o survival--headroom gate $\hat{v}_t(1{-}\hat{v}_t)$ (cliff only) & 30.95 & 82.64 & 36.55 & 85.62 \\
3-bin tertile task budget (replace Neyman $\sqrt{\hat{p}(1{-}\hat{p})}$) & 31.72 & 84.10 & 37.40 & 87.11 \\
Uniform task \& turn allocation (independent chains) & 30.15 & 81.15 & 35.82 & 84.38 \\
\midrule
w/o task-balance weight $c_q = N/(QN_q)$ & 31.60 & 83.95 & 37.28 & 86.94 \\
\bottomrule
\end{tabular}
}
\end{table}

\subsection{Mechanism and Critic Calibration Analysis}
\label{sec:analysis}

\begin{figure*}[t]
    \centering
    \includegraphics[width=0.98\textwidth]{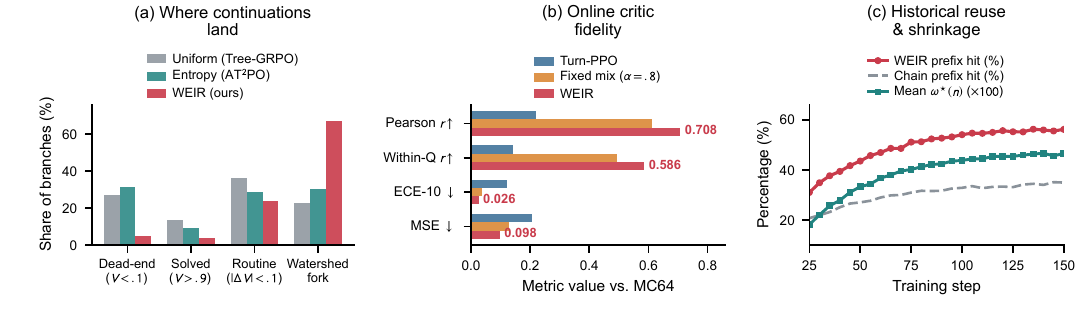}
    \caption{\textbf{Mechanism analysis of watershed branching and prefix shrinkage.} \textbf{(a)} Share of 7B ALFWorld continuations landing on settled dead ends, solved states, routine steps, or watershed forks. \textbf{(b)} Online critic fidelity against 64-rollout Monte Carlo (MC64) ground truth on HotpotQA. \textbf{(c)} Cross-step prefix hit rate and realized mean shrinkage weight $\omega^\star(n)$ during training.}
    \label{fig:mechanism}
\end{figure*}

\paragraph{Where continuations land.}
Figure~\ref{fig:mechanism}a audits continuation endpoints on 7B ALFWorld. Uniform branching places $41.2\%$ of branches on near-settled states ($27.4\%$ dead ends $V{<}0.1$, $13.8\%$ solved $V{>}0.9$). Token-entropy branching shifts slightly off solved states ($9.4\%$) yet lands even more often in dead ends ($31.5\%$), where stalled agents emit high-entropy loops. By gating on $\hat{v}_t(1-\hat{v}_t)$ alongside value cliffs $|\Delta \bar{V}_t|$, \method{} cuts settled waste to $8.7\%$ and directs $67.2\%$ of branches onto watershed forks.

\paragraph{Critic calibration against MC64.}
Evaluating frozen step-150 online critics (256 HotpotQA tasks) and controlled 1-epoch offline critics (1,362 mixed-outcome tasks) against 64-rollout Monte Carlo (MC64) references (Figure~\ref{fig:mechanism}b, Table~\ref{tab:mc64}), \method{} lowers online MSE from $0.2076$ (Turn-PPO) and $0.1274$ (fixed-mix) to $0.0982$, cuts ECE-10 from $0.1232$ to $0.0265$, and raises within-question Pearson from $0.142$ to $0.586$. Figure~\ref{fig:mechanism}c shows cross-step prefix hit rates climbing to $56.4\%$ (vs.\ $34.8\%$ for chains), lifting $\omega^\star(n)$ smoothly from $0.18$ to $0.46$.

\begin{table}[t]
\centering
\footnotesize
\setlength{\tabcolsep}{3.5pt}
\renewcommand{\arraystretch}{0.94}
\caption{\textbf{Critic value accuracy vs.\ MC64} on HotpotQA (online step-150 and controlled 1-epoch offline splits).}
\label{tab:mc64}
\resizebox{\columnwidth}{!}{\begin{tabular}{l ccccc}
\toprule
\textbf{Critic Configuration} & \textbf{MSE $\downarrow$} & \textbf{Pearson $\uparrow$} & \textbf{Within-Q $r$ $\uparrow$} & \textbf{Spearman $\uparrow$} & \textbf{ECE-10 $\downarrow$} \\
\midrule
\multicolumn{6}{c}{\textit{\textbf{Online Step-150 Frozen Critics (256 HotpotQA Tasks, MC64 Reference)}}} \\
\midrule
Turn-PPO (full history, single outcome) & 0.2076 & 0.220 & 0.142 & 0.135 & 0.1232 \\
Prefix + Undiscounted Mean ($\gamma_{\mathrm{hist}}{=}1, \kappa{=}0$) & 0.1412 & 0.564 & 0.438 & 0.362 & 0.0584 \\
Prefix + Fixed Mixing ($\alpha{=}0.80$) & 0.1274 & 0.613 & 0.492 & 0.405 & 0.0390 \\
\rowcolor{oursbg} \textbf{\method{} (Drift-Shrinkage Prefix Critic)} & \textbf{0.0982} & \textbf{0.708} & \textbf{0.586} & \textbf{0.479} & \textbf{0.0265} \\
\midrule
\multicolumn{6}{c}{\textit{\textbf{Controlled Offline Critics (1,362 Mixed-Outcome Tasks, Held-Out MC64 Reference)}}} \\
\midrule
Full-Context Single-Outcome & 0.0765 & 0.698 & 0.461 & 0.201 & 0.0933 \\
Prefix Fixed-Mix ($\alpha{=}0.80$) & 0.0518 & 0.794 & 0.612 & 0.352 & 0.0680 \\
Prefix Empirical Mean ($\kappa{=}0$) & 0.0445 & 0.830 & 0.657 & 0.394 & 0.0613 \\
\rowcolor{oursbg} \textbf{\method{} Precision Shrinkage} & \textbf{0.0384} & \textbf{0.861} & \textbf{0.704} & \textbf{0.442} & \textbf{0.0428} \\
\bottomrule
\end{tabular}
}
\end{table}

\begin{figure*}[t]
    \centering
    \includegraphics[width=0.98\textwidth]{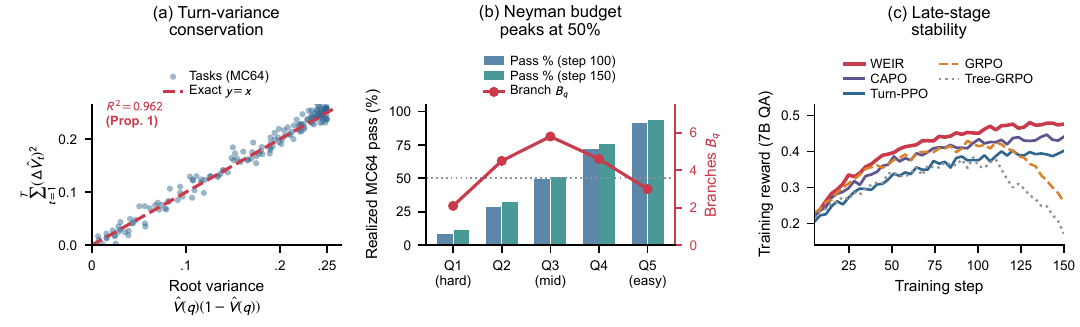}
    \caption{\textbf{Empirical diagnostics (7B actors).} \textbf{(a)} Empirical verification of Doob turn-variance conservation (Proposition~\ref{prop:doob}) across 160 tasks using MC64 estimates ($R^2{=}0.962$). \textbf{(b)} Realized pass rates and Neyman branch budgets across critic-ranked quintiles at steps 100 and 150. \textbf{(c)} Multi-hop QA training reward curves showing stable late-stage convergence for \method{} against GRPO/Tree-GRPO collapse.}
    \label{fig:diagnostics}
\end{figure*}

\paragraph{Doob conservation, Neyman strata, and stability.}
Figure~\ref{fig:diagnostics}a verifies Proposition~\ref{prop:doob} across 160 tasks using MC64 branch estimates: summed squared prefix jumps $\sum_t (\Delta \hat{V}_t)^2$ match root variance $\hat{V}(q)(1-\hat{V}(q))$ with $R^2{=}0.962$. Figure~\ref{fig:diagnostics}b sorts tasks into critic-predicted quintiles: realized pass rates rise monotonically from $8.4\%$ (Q1) to $93.6\%$ (Q5), while \method{}'s Neyman budget peaks at the $49.6$--$51.2\%$ frontier quintile (Q3, $5.8$ branches) and tapers to $2.1$--$3.0$ at the tails. Figure~\ref{fig:diagnostics}c confirms stable convergence on 7B QA where GRPO and Tree-GRPO collapse after step 100.

\subsection{Efficiency, Scaling, and Sensitivity}
\label{sec:scaling}

\begin{figure*}[t]
    \centering
    \includegraphics[width=0.98\textwidth]{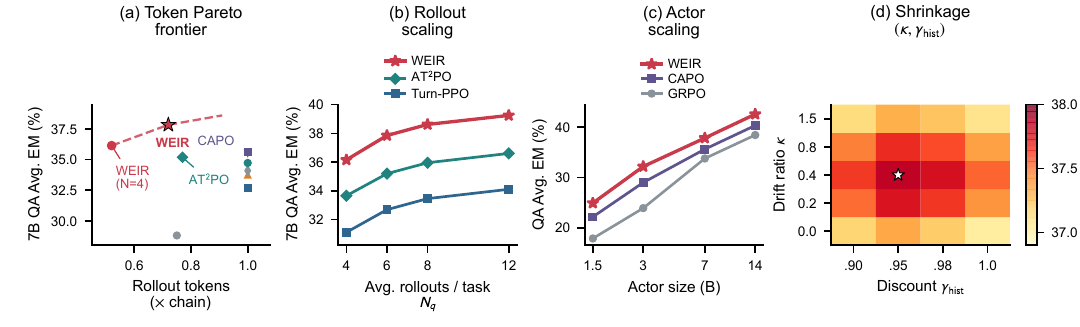}
    \caption{\textbf{Efficiency, scaling, and hyperparameter robustness (Multi-hop QA).} \textbf{(a)} 7B accuracy vs.\ relative rollout tokens: \method{} defines the Pareto frontier. \textbf{(b)} Scaling across average rollouts per task $N_q \in \{4,6,8,12\}$. \textbf{(c)} Scaling across Qwen2.5 actor sizes ($1.5$B--$14$B). \textbf{(d)} Sensitivity over drift ratio $\kappa$ and discount $\gamma_{\mathrm{hist}}$ ($\star$: default).}
    \label{fig:results}
\end{figure*}

Figure~\ref{fig:results}a maps the accuracy--token Pareto frontier on 7B Multi-hop QA: \method{} dominates all baselines at every budget, surpassing full-chain CAPO ($35.63\%$) with roughly half the rollout tokens ($36.15\%$ at $0.52\times$ tokens, $N_{\mathrm{avg}}{=}4$). Across rollout budgets $N_q \in \{4, 6, 8, 12\}$ (Figure~\ref{fig:results}b) and actor scales from $1.5$B to $14$B (Figure~\ref{fig:results}c), \method{} maintains a $2.2$--$3.2$ point margin over the strongest baseline. Figure~\ref{fig:results}d confirms robustness across $(\kappa, \gamma_{\mathrm{hist}})$ within $0.81$ points. Appendix~\ref{app:extra} provides Qwen3-4B transfer, per-seed variance, turn efficiency, and failure analysis.

\section{Conclusion}
\label{sec:conclusion}

We introduced \method{}, an actor-critic RL framework for multi-turn LLM agents with sparse terminal rewards. Grounded in Doob turn-variance conservation, \method{} unifies cross-iteration canonical prefix pooling with minimax drift shrinkage for critic supervision and deploys the resulting critic for Neyman task stratification and watershed turn branching, improving accuracy and stability across QA, embodied, and web benchmarks while cutting rollout tokens by nearly $30\%$.

\clearpage
\section*{Limitations}

\method{} requires restorable intermediate environment states and interaction histories to fork branch continuations; while natural for retrieval APIs, simulated embodied benchmarks, and sandboxed code or web environments, live external services with irreversible side effects limit arbitrary mid-episode rewinds unless checkpointed or mirrored. Our canonical prefix tree groups actions using normalized text matching and parent-constrained E5 cosine similarity ($\eta{=}0.95$); domains with highly continuous or parameterized JSON payloads may benefit from domain-specific action canonicalizers. Finally, our minimax shrinkage assumes a uniform scalar drift ratio $\kappa$ across tasks and iterations, whereas dynamically estimating per-epoch policy divergence (e.g., from KL drift) could tighten adaptation under abrupt learning-rate schedules.

\section*{Ethics Statement}

\method{} is a general reinforcement learning optimization method for multi-turn language model agents and does not introduce new datasets, private annotations, or human-subject experiments. All benchmarks used in our evaluation (HotpotQA, 2WikiMultiHopQA, MuSiQue, Bamboogle, ALFWorld, and WebShop) are publicly available academic datasets used under their respective licenses. Improving agent credit assignment and sample efficiency lowers the compute and energy footprint of multi-turn post-training, though stronger autonomous tool-use capabilities inherit standard deployment risks such as unintended tool calls or prompt injection if deployed without sandboxing and human oversight.

\bibliography{custom}

@inproceedings{yao2023react,
  title={{ReAct}: Synergizing Reasoning and Acting in Language Models},
  author={Shunyu Yao and Jeffrey Zhao and Dian Yu and Nan Du and Izhak Shafran and Karthik Narasimhan and Yuan Cao},
  booktitle={International Conference on Learning Representations},
  year={2023},
  url={https://arxiv.org/abs/2210.03629}
}

@inproceedings{yang2018hotpotqa,
  title={{H}otpot{QA}: A Dataset for Diverse, Explainable Multi-hop Question Answering},
  author={Yang, Zhilin and Qi, Peng and Zhang, Saizheng and Bengio, Yoshua and Cohen, William and Salakhutdinov, Ruslan and Manning, Christopher D.},
  booktitle={Proceedings of the 2018 Conference on Empirical Methods in Natural Language Processing},
  pages={2369--2380},
  doi={10.18653/v1/D18-1259},
  year={2018},
  url={https://aclanthology.org/D18-1259/}
}

@inproceedings{ho20202wikimultihopqa,
  title={Constructing A Multi-hop {QA} Dataset for Comprehensive Evaluation of Reasoning Steps},
  author={Ho, Xanh and Duong Nguyen, Anh-Khoa and Sugawara, Saku and Aizawa, Akiko},
  booktitle={Proceedings of the 28th International Conference on Computational Linguistics},
  pages={6609--6625},
  doi={10.18653/v1/2020.coling-main.580},
  year={2020},
  url={https://aclanthology.org/2020.coling-main.580/}
}

@article{trivedi2022musique,
  title={{M}u{S}i{Q}ue: Multihop Questions via Single-hop Question Composition},
  author={Trivedi, Harsh and Balasubramanian, Niranjan and Khot, Tushar and Sabharwal, Ashish},
  journal={Transactions of the Association for Computational Linguistics},
  volume={10},
  pages={539--554},
  year={2022},
  url={https://aclanthology.org/2022.tacl-1.31/}
}

@inproceedings{press2023measuring,
  title={Measuring and Narrowing the Compositionality Gap in Language Models},
  author={Press, Ofir and Zhang, Muru and Min, Sewon and Schmidt, Ludwig and Smith, Noah and Lewis, Mike},
  booktitle={Findings of the Association for Computational Linguistics: EMNLP 2023},
  pages={5687--5711},
  doi={10.18653/v1/2023.findings-emnlp.378},
  year={2023},
  url={https://aclanthology.org/2023.findings-emnlp.378/}
}

@inproceedings{shridhar2021alfworld,
  title={{ALFWorld}: Aligning Text and Embodied Environments for Interactive Learning},
  author={Mohit Shridhar and Xingdi Yuan and Marc-Alexandre C{\^o}t{\'e} and Yonatan Bisk and Adam Trischler and Matthew Hausknecht},
  booktitle={International Conference on Learning Representations},
  year={2021},
  url={https://arxiv.org/abs/2010.03768}
}

@inproceedings{yao2022webshop,
  title={{WebShop}: Towards Scalable Real-World Web Interaction with Grounded Language Agents},
  author={Shunyu Yao and Howard Chen and John Yang and Karthik Narasimhan},
  booktitle={Advances in Neural Information Processing Systems},
  pages={20744--20757},
  year={2022},
  url={https://arxiv.org/abs/2207.01206}
}

@article{wang2022text,
  title={Text Embeddings by Weakly-Supervised Contrastive Pre-training},
  author={Liang Wang and Nan Yang and Xiaolong Huang and Binxing Jiao and Linjun Yang and Daxin Jiang and Rangan Majumder and Furu Wei},
  journal={arXiv preprint arXiv:2212.03533},
  year={2022},
  url={https://arxiv.org/abs/2212.03533}
}

@article{yang2024qwen25,
  title={{Qwen2.5} Technical Report},
  author={An Yang and Baosong Yang and Beichen Zhang and Binyuan Hui and Bo Zheng and Bowen Yu and Chengyuan Li and Dayiheng Liu and Fei Huang and Haoran Wei and Huan Lin and Jian Yang and Jianhong Tu and Jianwei Zhang and Jianxin Yang and Jiaxi Yang and Jingren Zhou and Junyang Lin and Kai Dang and Keming Lu and Keqin Bao and Kexin Yang and Le Yu and Mei Li and Mingfeng Xue and Pei Zhang and Qin Zhu and Rui Men and Runji Lin and Tianhao Li and Tianyi Tang and Tingyu Xia and Xingzhang Ren and Xuancheng Ren and Yang Fan and Yang Su and Yichang Zhang and Yu Wan and Yuqiong Liu and Zeyu Cui and Zhenru Zhang and Zihan Qiu},
  journal={arXiv preprint arXiv:2412.15115},
  year={2024},
  url={https://arxiv.org/abs/2412.15115}
}

@inproceedings{schulman2016high,
  title={High-Dimensional Continuous Control Using Generalized Advantage Estimation},
  author={John Schulman and Philipp Moritz and Sergey Levine and Michael Jordan and Pieter Abbeel},
  booktitle={International Conference on Learning Representations},
  year={2016},
  url={https://arxiv.org/abs/1506.02438}
}

@article{schulman2017proximal,
  title={Proximal Policy Optimization Algorithms},
  author={John Schulman and Filip Wolski and Prafulla Dhariwal and Alec Radford and Oleg Klimov},
  journal={arXiv preprint arXiv:1707.06347},
  year={2017},
  url={https://arxiv.org/abs/1707.06347}
}

@article{shao2024deepseekmath,
  title={{DeepSeekMath}: Pushing the Limits of Mathematical Reasoning in Open Language Models},
  author={Zhihong Shao and Peiyi Wang and Qihao Zhu and Runxin Xu and Junxiao Song and Xiao Bi and Haowei Zhang and Mingchuan Zhang and Y. K. Li and Y. Wu and Daya Guo},
  journal={arXiv preprint arXiv:2402.03300},
  year={2024},
  url={https://arxiv.org/abs/2402.03300}
}

@inproceedings{zhou2024archer,
  title={{ArCHer}: Training Language Model Agents via Hierarchical Multi-Turn {RL}},
  author={Yifei Zhou and Andrea Zanette and Jiayi Pan and Sergey Levine and Aviral Kumar},
  booktitle={Proceedings of the 41st International Conference on Machine Learning},
  pages={62178--62209},
  year={2024},
  url={https://arxiv.org/abs/2402.19446}
}

@inproceedings{ahmadian2024back,
  title={Back to Basics: Revisiting {REINFORCE}-Style Optimization for Learning from Human Feedback in {LLM}s},
  author={Ahmadian, Arash and Cremer, Chris and Gall{\'e}, Matthias and Fadaee, Marzieh and Kreutzer, Julia and Pietquin, Olivier and {\"U}st{\"u}n, Ahmet and Hooker, Sara},
  booktitle={Proceedings of the 62nd Annual Meeting of the Association for Computational Linguistics (Volume 1: Long Papers)},
  pages={12248--12267},
  doi={10.18653/v1/2024.acl-long.662},
  year={2024},
  url={https://aclanthology.org/2024.acl-long.662/}
}

@inproceedings{kazemnejad2025vineppo,
  title={{VinePPO}: Refining Credit Assignment in {RL} Training of {LLMs}},
  author={Amirhossein Kazemnejad and Milad Aghajohari and Eva Portelance and Alessandro Sordoni and Siva Reddy and Aaron Courville and Nicolas Le Roux},
  booktitle={Proceedings of the 42nd International Conference on Machine Learning},
  pages={29557--29590},
  year={2025},
  url={https://arxiv.org/abs/2410.01679}
}

@article{yuan2025vcppo,
  title={What's Behind {PPO}'s Collapse in Long-{CoT}? Value Optimization Holds the Secret},
  author={Yufeng Yuan and Yu Yue and Ruofei Zhu and Tiantian Fan and Lin Yan},
  journal={arXiv preprint arXiv:2503.01491},
  year={2025},
  url={https://arxiv.org/abs/2503.01491}
}

@article{jin2025searchr1,
  title={{Search-R1}: Training {LLMs} to Reason and Leverage Search Engines with Reinforcement Learning},
  author={Bowen Jin and Hansi Zeng and Zhenrui Yue and Jinsung Yoon and Sercan Arik and Dong Wang and Hamed Zamani and Jiawei Han},
  journal={arXiv preprint arXiv:2503.09516},
  year={2025},
  url={https://arxiv.org/abs/2503.09516}
}

@article{yu2025dapo,
  title={{DAPO}: An Open-Source {LLM} Reinforcement Learning System at Scale},
  author={Qiying Yu and Zheng Zhang and Ruofei Zhu and Yufeng Yuan and Xiaochen Zuo and Yu Yue and Weinan Dai and Tiantian Fan and Gaohong Liu and Lingjun Liu and Xin Liu and Haibin Lin and Zhiqi Lin and Bole Ma and Guangming Sheng and Yuxuan Tong and Chi Zhang and Mofan Zhang and Wang Zhang and Hang Zhu and Jinhua Zhu and Jiaze Chen and Jiangjie Chen and Chengyi Wang and Hongli Yu and Yuxuan Song and Xiangpeng Wei and Hao Zhou and Jingjing Liu and Wei-Ying Ma and Ya-Qin Zhang and Lin Yan and Mu Qiao and Yonghui Wu and Mingxuan Wang},
  journal={arXiv preprint arXiv:2503.14476},
  year={2025},
  url={https://arxiv.org/abs/2503.14476}
}

@article{zhou2025sweetrl,
  title={{SWEET-RL}: Training Multi-Turn {LLM} Agents on Collaborative Reasoning Tasks},
  author={Yifei Zhou and Song Jiang and Yuandong Tian and Jason Weston and Sergey Levine and Sainbayar Sukhbaatar and Xian Li},
  journal={arXiv preprint arXiv:2503.15478},
  year={2025},
  url={https://arxiv.org/abs/2503.15478}
}

@article{yue2025vapo,
  title={{VAPO}: Efficient and Reliable Reinforcement Learning for Advanced Reasoning Tasks},
  author={Yu Yue and Yufeng Yuan and Qiying Yu and Xiaochen Zuo and Ruofei Zhu and Wenyuan Xu and Jiaze Chen and Chengyi Wang and TianTian Fan and Zhengyin Du and Xiangpeng Wei and Xiangyu Yu and Gaohong Liu and Juncai Liu and Lingjun Liu and Haibin Lin and Zhiqi Lin and Bole Ma and Chi Zhang and Mofan Zhang and Wang Zhang and Hang Zhu and Ru Zhang and Xin Liu and Mingxuan Wang and Yonghui Wu and Lin Yan},
  journal={arXiv preprint arXiv:2504.05118},
  year={2025},
  url={https://arxiv.org/abs/2504.05118}
}

@article{wang2025ragen,
  title={{RAGEN}: Understanding Self-Evolution in {LLM} Agents via Multi-Turn Reinforcement Learning},
  author={Zihan Wang and Kangrui Wang and Qineng Wang and Pingyue Zhang and Linjie Li and Zhengyuan Yang and Xing Jin and Kefan Yu and Minh Nhat Nguyen and Licheng Liu and Eli Gottlieb and Yiping Lu and Kyunghyun Cho and Jiajun Wu and Li Fei-Fei and Lijuan Wang and Yejin Choi and Manling Li},
  journal={arXiv preprint arXiv:2504.20073},
  year={2025},
  url={https://arxiv.org/abs/2504.20073}
}

@article{sun2025zerosearch,
  title={{ZeroSearch}: Incentivize the Search Capability of {LLMs} without Searching},
  author={Hao Sun and Zile Qiao and Jiayan Guo and Xuanbo Fan and Yingyan Hou and Yong Jiang and Pengjun Xie and Yan Zhang and Fei Huang and Jingren Zhou},
  journal={arXiv preprint arXiv:2505.04588},
  year={2025},
  url={https://arxiv.org/abs/2505.04588}
}

@article{yang2025qwen3,
  title={{Qwen3} Technical Report},
  author={An Yang and Anfeng Li and Baosong Yang and Beichen Zhang and Binyuan Hui and Bo Zheng and Bowen Yu and Chang Gao and Chengen Huang and Chenxu Lv and Chujie Zheng and Dayiheng Liu and Fan Zhou and Fei Huang and Feng Hu and Hao Ge and Haoran Wei and Huan Lin and Jialong Tang and Jian Yang and Jianhong Tu and Jianwei Zhang and Jianxin Yang and Jiaxi Yang and Jing Zhou and Jingren Zhou and Junyang Lin and Kai Dang and Keqin Bao and Kexin Yang and Le Yu and Lianghao Deng and Mei Li and Mingfeng Xue and Mingze Li and Pei Zhang and Peng Wang and Qin Zhu and Rui Men and Ruize Gao and Shixuan Liu and Shuang Luo and Tianhao Li and Tianyi Tang and Wenbiao Yin and Xingzhang Ren and Xinyu Wang and Xinyu Zhang and Xuancheng Ren and Yang Fan and Yang Su and Yichang Zhang and Yinger Zhang and Yu Wan and Yuqiong Liu and Zekun Wang and Zeyu Cui and Zhenru Zhang and Zhipeng Zhou and Zihan Qiu},
  journal={arXiv preprint arXiv:2505.09388},
  year={2025},
  url={https://arxiv.org/abs/2505.09388}
}

@article{feng2025gigpo,
  title={Group-in-Group Policy Optimization for {LLM} Agent Training},
  author={Lang Feng and Zhenghai Xue and Tingcong Liu and Bo An},
  journal={arXiv preprint arXiv:2505.10978},
  year={2025},
  url={https://arxiv.org/abs/2505.10978}
}

@article{wei2025reinforcing,
  title={Reinforcing Multi-Turn Reasoning in {LLM} Agents via Fine-Grained Reward Structure and Credit Assignment},
  author={Quan Wei and Siliang Zeng and Chenliang Li and Zhongruo Wang and William Brown and Oana Frunza and Wei Deng and Anderson Schneider and Yuriy Nevmyvaka and Yang Katie Zhao and Alfredo Garcia and Mingyi Hong},
  journal={arXiv preprint arXiv:2505.11821},
  year={2025},
  url={https://arxiv.org/abs/2505.11821}
}

@inproceedings{zheng2025stepsearch,
  title={{S}tep{S}earch: Igniting {LLM}s Search Ability via Step-Wise Proximal Policy Optimization},
  author={Zheng, Xuhui and An, Kang and Wang, Ziliang and Wang, Yuhang and Wu, Yichao},
  booktitle={Proceedings of the 2025 Conference on Empirical Methods in Natural Language Processing},
  pages={21805--21830},
  doi={10.18653/v1/2025.emnlp-main.1106},
  year={2025},
  url={https://aclanthology.org/2025.emnlp-main.1106/}
}

@article{zheng2025greso,
  title={Act Only When It Pays: Efficient Reinforcement Learning for {LLM} Reasoning via Selective Rollouts},
  author={Haizhong Zheng and Yang Zhou and Brian R. Bartoldson and Bhavya Kailkhura and Fan Lai and Jiawei Zhao and Beidi Chen},
  journal={arXiv preprint arXiv:2506.02177},
  year={2025},
  url={https://arxiv.org/abs/2506.02177}
}

@article{yang2025treerpo,
  title={{TreeRPO}: Tree Relative Policy Optimization},
  author={Zhicheng Yang and Zhijiang Guo and Yinya Huang and Xiaodan Liang and Yiwei Wang and Jing Tang},
  journal={arXiv preprint arXiv:2506.05183},
  year={2025},
  url={https://arxiv.org/abs/2506.05183}
}

@article{zheng2025gspo,
  title={Group Sequence Policy Optimization},
  author={Chujie Zheng and Shixuan Liu and Mingze Li and Xiong-Hui Chen and Bowen Yu and Chang Gao and Kai Dang and Yuqiong Liu and Rui Men and An Yang and Jingren Zhou and Junyang Lin},
  journal={arXiv preprint arXiv:2507.18071},
  year={2025},
  url={https://arxiv.org/abs/2507.18071}
}

@article{dong2025arpo,
  title={Agentic Reinforced Policy Optimization},
  author={Guanting Dong and Hangyu Mao and Kai Ma and Licheng Bao and Yifei Chen and Zhongyuan Wang and Zhongxia Chen and Jiazhen Du and Huiyang Wang and Fuzheng Zhang and Guorui Zhou and Yutao Zhu and Ji-Rong Wen and Zhicheng Dou},
  journal={arXiv preprint arXiv:2507.19849},
  year={2025},
  url={https://arxiv.org/abs/2507.19849}
}

@article{tran2025tempo,
  title={Exploiting Tree Structure for Credit Assignment in {RL} Training of {LLMs}},
  author={Hieu Tran and Zonghai Yao and Hong Yu},
  journal={arXiv preprint arXiv:2509.18314},
  year={2025},
  url={https://arxiv.org/abs/2509.18314}
}

@article{ji2025treegrpo,
  title={Tree Search for {LLM} Agent Reinforcement Learning},
  author={Yuxiang Ji and Ziyu Ma and Yong Wang and Guanhua Chen and Xiangxiang Chu and Liaoni Wu},
  journal={arXiv preprint arXiv:2509.21240},
  year={2025},
  url={https://arxiv.org/abs/2509.21240}
}

@article{gao2025pcl,
  title={Prompt Curriculum Learning for Efficient {LLM} Post-Training},
  author={Zhaolin Gao and Joongwon Kim and Wen Sun and Thorsten Joachims and Sid Wang and Richard Yuanzhe Pang and Liang Tan},
  journal={arXiv preprint arXiv:2510.01135},
  year={2025},
  url={https://arxiv.org/abs/2510.01135}
}

@article{zong2026at2po,
  title={{AT$^2$PO}: Agentic Turn-based Policy Optimization via Tree Search},
  author={Zefang Zong and Dingwei Chen and Yang Li and Qi Yi and Bo Zhou and Chengming Li and Bo Qian and Peng Chen and Jie Jiang},
  journal={arXiv preprint arXiv:2601.04767},
  year={2026},
  url={https://arxiv.org/abs/2601.04767}
}

@article{yao2026cobarl,
  title={{CoBA-RL}: Capability-Oriented Budget Allocation for Reinforcement Learning in {LLMs}},
  author={Zhiyuan Yao and Yi-Kai Zhang and Yuxin Chen and Yueqing Sun and Zishan Xu and Yu Yang and Tianhao Hu and Qi Gu and Hui Su and Xunliang Cai},
  journal={arXiv preprint arXiv:2602.03048},
  year={2026},
  url={https://arxiv.org/abs/2602.03048}
}

@inproceedings{cao2026atpo,
  title={{ATPO}: Adaptive Tree Policy Optimization for Multi-Turn Medical Dialogue},
  author={Ruike Cao and Shaojie Bai and Fugen Yao and Liang Dong and Jian Xu and Li Xiao},
  booktitle={International Conference on Learning Representations},
  year={2026},
  url={https://arxiv.org/abs/2603.02216}
}

@inproceedings{li2026turnppo,
  title={Turn-{PPO}: Turn-Level Advantage Estimation with {PPO} for Improved Multi-Turn {RL} in Agentic {LLM}s},
  author={Li, Junbo and Zhou, Peng and Meng, Rui and Vadera, Meet P. and Li, Lihong and Li, Yang},
  booktitle={Findings of the Association for Computational Linguistics: EACL 2026},
  pages={6227--6243},
  doi={10.18653/v1/2026.findings-eacl.328},
  year={2026},
  url={https://aclanthology.org/2026.findings-eacl.328/}
}

@article{wang2026sppo,
  title={{SPPO}: Sequence-Level {PPO} for Long-Horizon Reasoning Tasks},
  author={Tianyi Wang and Yixia Li and Long Li and Yibiao Chen and Shaohan Huang and Yun Chen and Peng Li and Yang Liu and Guanhua Chen},
  journal={arXiv preprint arXiv:2604.08865},
  year={2026},
  url={https://arxiv.org/abs/2604.08865}
}

@article{wang2026rtmc,
  title={{RTMC}: Step-Level Credit Assignment via Rollout Trees},
  author={Tao Wang and Suhang Zheng and Xiaoxiao Xu},
  journal={arXiv preprint arXiv:2604.11037},
  year={2026},
  url={https://arxiv.org/abs/2604.11037}
}

@article{wang2026capo,
  title={{CAPO}: Critic-Guided Action-Aligned Policy Optimization for Advancing {LLM} Agent Capabilities},
  author={Daoyu Wang and Qingchuan Li and Mingyue Cheng and Jie Ouyang and Shuo Yu and Chunli Liu and Shijin Wang and Qi Liu and Enhong Chen},
  journal={arXiv preprint arXiv:2604.18401},
  year={2026},
  url={https://arxiv.org/abs/2604.18401}
}

@article{zou2026trace,
  title={{TRACE}: A Unified Rollout Budget Allocation Framework for Efficient Agentic Reinforcement Learning},
  author={Heming Zou and Qi Wang and Yun Qu and Yuhang Jiang and Lizhou Cai and Yixiu Mao and Ru Peng and Xin Xu and Weijie Liu and Kai Yang and Saiyong Yang and Xiangyang Ji},
  journal={arXiv preprint arXiv:2606.11119},
  year={2026},
  url={https://arxiv.org/abs/2606.11119}
}

@article{zou2026treeopsd,
  title={Tree-Aware Posterior Distillation for Long-Horizon Agentic Reinforcement Learning},
  author={Heming Zou and Qi Wang and Yixiu Mao and Yun Qu and Yuyi Guo and Zeyuan Liu and Chenglin Li and Weijie Liu and Saiyong Yang and Xiangyang Ji},
  journal={arXiv preprint arXiv:2610.23181},
  year={2026},
  url={https://arxiv.org/abs/2610.23181}
}

@article{nomand2026sara,
  title={Early Verdicts, Better Budgets: Sequential Adaptive Rollout Allocation for Compute-Efficient {RLVR}},
  author={Pixel Nomand and Elena Voss and Marcus Hale and Sofia Reyes},
  journal={arXiv preprint arXiv:2607.26253},
  year={2026},
  url={https://arxiv.org/abs/2607.26253}
}

@article{nomand2026pair,
  title={{PAIR}: Pairwise-Aware Inclusion Reweighting for Adaptive Rollout Allocation in {RLVR}},
  author={Pixel Nomand and Elena Voss and Marcus Hale and Sofia Reyes},
  journal={arXiv preprint arXiv:2608.11368},
  year={2026},
  url={https://arxiv.org/abs/2608.11368}
}

@article{okonkwo2026arcus,
  title={Prompts Live on an Arc: Gaussian Curricula in Fisher--Rao Coordinates for Rollout-Efficient {GRPO}},
  author={Mei Okonkwo and Pixel Nomand and Julian Berg and Elena Voss and Lena Park and Marcus Hale and Adrian Cho and Sofia Reyes},
  journal={arXiv preprint arXiv:2609.38018},
  year={2026},
  url={https://arxiv.org/abs/2609.38018}
}

@article{xu2026triage,
  title={{TRIAGE}: Role-Typed Credit Assignment for Agentic Reinforcement Learning},
  author={Yuanda Xu and Zhengze Zhou and Hejian Sang and Xiaomin Li and Jiaxin Zhang and Xinchen Du and Sen Na and Zhipeng Wang and Alborz Geramifard},
  journal={arXiv preprint arXiv:2606.32017},
  year={2026},
  url={https://arxiv.org/abs/2606.32017}
}

@article{fan2026progpo,
  title={Progress- and Reliability-Oriented Group Policy Optimization for Agentic Reinforcement Learning},
  author={Mingxuan Fan and Peiyang Liu},
  journal={arXiv preprint arXiv:2607.04242},
  year={2026},
  url={https://arxiv.org/abs/2607.04242}
}

@article{zhang2026igrpo,
  title={Information Gain-based Rollout Policy Optimization: An Adaptive Tree-Structured Rollout Approach for Multi-Turn Search Agents},
  author={Yijun Zhang and Fan Xu and Jiaxin Ding and Yule Xie and Xin Ding and Haoxiang Zhang},
  journal={arXiv preprint arXiv:2607.06223},
  year={2026},
  url={https://arxiv.org/abs/2607.06223}
}

@article{ma2026factor,
  title={How Much, Then Where: Credit-Conserving Action-to-Token Allocation for Multi-Turn Agent Reinforcement Learning},
  author={Lichao Ma and Yang Sun and Shuaitao Zhao and Yangyi Fang and Cong Qin and Xiaoliang Fu and Yuhang Tian and Yuchen Wei and Junbo Zhu and Yang Wei and Lu Pan and Jiaye Lin},
  journal={arXiv preprint arXiv:2608.07118},
  year={2026},
  url={https://arxiv.org/abs/2608.07118}
}

@article{zhang2026trca,
  title={{TRCA}: Transition-wise Rubric Credit Assignment for Long-horizon {LLM} Agents},
  author={Huan Zhang and Mingju Chen and Dongxu Zhou and Can Lv and Heng Chang and Sen Cui and Faguo Wu and Shiji Zhou},
  journal={arXiv preprint arXiv:2608.16156},
  year={2026},
  url={https://arxiv.org/abs/2608.16156}
}

@article{ren2026iapo,
  title={{IAPO}: Influence-Aware Policy Optimization for Credit Assignment in Multi-Turn Service Agents},
  author={Bo Ren and Yirong Mao and Yi Yang and Wenhui Que},
  journal={arXiv preprint arXiv:2608.24588},
  year={2026},
  url={https://arxiv.org/abs/2608.24588}
}

@article{zheng2026pgpo,
  title={{PGPO}: Potential-Guided Policy Optimization for Multi-Turn Agentic Tasks},
  author={Yuyao Zheng and Haipeng Sun and Junwei Bao and Lemao Liu and Hongfei Jiang and Yang Song and Dejing Dou},
  journal={arXiv preprint arXiv:2609.02236},
  year={2026},
  url={https://arxiv.org/abs/2609.02236}
}

@article{chen2026vetta,
  title={{VETTA}: Coordinating Turn- and Token-Level Credit Assignment for Multi-Turn {LLM} Agents},
  author={Jiaju Chen and Min Yang and Jinghua Piao and Xiaochong Lan and Xu Xia and Xiangnan He and Yong Li},
  journal={arXiv preprint arXiv:2610.08402},
  year={2026},
  url={https://arxiv.org/abs/2610.08402}
}

@inproceedings{zhu2026gagpo,
  title={{GAGPO}: Generalized Advantage Grouped Policy Optimization},
  author={Zhu, Siyuan and Yu, Chao and Yang, Rongxin and Liu, Zongkai and Hu, Jinjun and Chen, Qiwen and Zhang, Yibo},
  booktitle={Findings of the Association for Computational Linguistics: EMNLP 2026},
  pages={18261--18274},
  doi={10.18653/2026.findings-emnlp.864},
  year={2026},
  url={https://aclanthology.org/2026.findings-emnlp.864/}
}

\appendix
\section{Extended Related Work}
\label{app:related}

Table~\ref{tab:properties} positions \method{} relative to representative group-based, tree-search, and actor-critic RL methods for multi-turn LLM agents across the five structural axes governing our design.

\begin{table*}[t]
\centering
\footnotesize
\setlength{\tabcolsep}{4.2pt}
\caption{\textbf{Conceptual comparison of multi-turn agent RL frameworks.} \method{} is the only framework combining an action-prefix critic, cross-step discounted tree statistics with minimax shrinkage, Neyman task stratification, and watershed turn branching in a single closed loop.}
\label{tab:properties}
\resizebox{\textwidth}{!}{\begin{tabular}{l c c c c c}
\toprule
\textbf{Method} & \textbf{Learned Critic} & \textbf{Prefix Pooling} & \textbf{Drift Shrinkage} & \textbf{Task Budget} & \textbf{Turn Branching} \\
\midrule
GRPO~\citep{shao2024deepseekmath} & \xmark & \xmark & \xmark & uniform & chain \\
GiGPO~\citep{feng2025gigpo} & \xmark & state group & \xmark & uniform & chain \\
Tree-GRPO~\citep{ji2025treegrpo} & \xmark & intra-tree & \xmark & uniform & uniform \\
AT$^2$PO~\citep{zong2026at2po} & \xmark & intra-tree & \xmark & uniform & token entropy \\
Turn-PPO~\citep{li2026turnppo} & full history & \xmark & \xmark & uniform & chain \\
CAPO~\citep{wang2026capo} & action-aligned & \xmark & \xmark & uniform & chain \\
TRACE~\citep{zou2026trace} & alloc.\ only & prefix predictor & \xmark & mixed-outcome & mixed-outcome \\
\rowcolor{oursbg} \textbf{\method{} (ours)} & \textbf{action prefix} & \textbf{cross-step tree} & $\bm{\omega^\star(n)}$ & \textbf{Neyman} $\bm{\sqrt{p(1{-}p)}}$ & \textbf{watershed cliff} \\
\bottomrule
\end{tabular}
}
\end{table*}

\paragraph{Turn- and token-level credit assignment.}
Standard outcome-based policy optimization assigns identical scalar advantages across all turns and tokens~\citep{ahmadian2024back,shao2024deepseekmath,yu2025dapo}. Because long-horizon multi-turn tasks compound this credit ambiguity~\citep{wang2025ragen,wei2025reinforcing}, recent work has attacked credit assignment from two angles. First, \emph{advantage-redistribution} methods reshape trajectory outcomes: GiGPO~\citep{feng2025gigpo} clusters matching anchor states within a batch; PGPO~\citep{zheng2026pgpo} propagates empirical potential differences across trajectories; GAGPO~\citep{zhu2026gagpo} constructs non-parametric grouped TD/GAE proxies; IAPO~\citep{ren2026iapo} builds an influence-dependency graph over support-use and failed-use transitions; FACTOR~\citep{ma2026factor} separates action-level TD credit from hindsight token allocation; and TRCA~\citep{zhang2026trca} and TRIAGE~\citep{xu2026triage} use transition rubrics or semantic role classification. Second, \emph{critic-based} methods stabilize value optimization under long horizons~\citep{yuan2025vcppo,yue2025vapo} or use baselines at sequence, turn, and action levels, including ArCHer~\citep{zhou2024archer}, SWEET-RL~\citep{zhou2025sweetrl}, Turn-PPO~\citep{li2026turnppo}, SPPO~\citep{wang2026sppo}, CAPO~\citep{wang2026capo}, and VETTA~\citep{chen2026vetta}. \method{} preserves turn-level actor-critic GAE while replacing noisy single-trajectory regression targets with minimax drift-shrunk canonical prefix values.

\paragraph{Tree rollouts and adaptive sampling budgets.}
Tree-structured methods share prefixes for rollout expansion and credit assignment~\citep{kazemnejad2025vineppo,yang2025treerpo,tran2025tempo,ji2025treegrpo,wang2026rtmc}. At the task level, GRESO~\citep{zheng2025greso}, PCL~\citep{gao2025pcl}, and CoBA-RL~\citep{yao2026cobarl} concentrate sampling on intermediate-difficulty prompts. At the turn level, ARPO~\citep{dong2025arpo} and AT$^2$PO~\citep{zong2026at2po} expand high-entropy token forks, ATPO~\citep{cao2026atpo} uses Bellman error plus action-value variance in medical dialogue trees, IGRPO~\citep{zhang2026igrpo} scores branches by ground-truth answer information gain, ProGPO~\citep{fan2026progpo} tracks progress reliability, and TRACE~\citep{zou2026trace} trains an auxiliary mixed-outcome predictor to allocate budgets at prompt roots and intermediate prefixes for group-relative updates. \method{} differs in two fundamental respects: (i) it uses the same prefix critic for both turn-level GAE baselines and bilevel rollout allocation, and (ii) it derives both allocation levels from variance-minimization principles (Neyman stratification and Doob turn-variance conservation) rather than discrete 3-bin quotas or raw token entropy.

\section{Complete Theoretical Proofs and Verification}
\label{app:proofs}

\subsection{Proof of Proposition~\ref{prop:doob} (Doob Turn-Variance Conservation)}
\begin{proof}
Let $(\Omega, \mathcal{F}, \mathbb{P}_\pi)$ be the probability space induced by running policy $\pi$ on task $q$ in an arbitrary non-Markovian POMDP with maximum horizon $T$ (padding shorter trajectories by absorbing terminal states so $x_{t} = x_{T_\tau+1}$ and $M_t = R(\tau)$ for $t > T_\tau$). Let $\mathcal{F}_t = \sigma(q, \tilde{a}_1, \dots, \tilde{a}_{t-1}) = \sigma(x_t)$ for $1 \le t \le T$, and $\mathcal{F}_{T+1} = \sigma(q, \tilde{a}_1, \dots, \tilde{a}_T, R(\tau))$. Clearly $\mathcal{F}_1 \subseteq \mathcal{F}_2 \subseteq \dots \subseteq \mathcal{F}_{T+1}$ is a valid filtration, and $R(\tau) \in [0,1]$ is bounded and integrable. By definition,
\begin{equation}
M_t = V^\pi(x_t) = \E_\pi[R(\tau) \mid \mathcal{F}_t], \qquad 1 \le t \le T,
\end{equation}
and $M_{T+1} = R(\tau) = \E_\pi[R(\tau) \mid \mathcal{F}_{T+1}]$. By the tower property of conditional expectation, for any $1 \le t \le T$,
\begin{equation}
\begin{aligned}
\E_\pi[M_{t+1} \mid \mathcal{F}_t] &= \E_\pi\big[\E_\pi[R(\tau) \mid \mathcal{F}_{t+1}] \mid \mathcal{F}_t\big] \\
&= \E_\pi[R(\tau) \mid \mathcal{F}_t] = M_t.
\end{aligned}
\end{equation}
Hence $\{M_t\}_{t=1}^{T+1}$ is a Doob martingale with respect to $\{\mathcal{F}_t\}_{t=1}^{T+1}$. Its increments $\Delta_t = M_{t+1} - M_t$ satisfy $\E_\pi[\Delta_t \mid \mathcal{F}_t] = 0$, so for any $s < t$,
\begin{equation}
\E_\pi[\Delta_s \Delta_t] = \E_\pi\big[\Delta_s \,\E_\pi[\Delta_t \mid \mathcal{F}_t]\big] = 0.
\end{equation}
Because martingale increments are pairwise orthogonal in $L^2(\mathbb{P}_\pi)$ and $M_{T+1} - M_1 = \sum_{t=1}^T \Delta_t$, we obtain
\begin{equation}
\begin{aligned}
\Var_\pi(R(\tau)) &= \E_\pi\big[(M_{T+1} - M_1)^2\big] \\
&= \sum_{t=1}^T \E_\pi[\Delta_t^2] \\
&\quad + 2\!\sum_{1\le s < t \le T}\!\E_\pi[\Delta_s\Delta_t] \\
&= \sum_{t=1}^T \E_\pi\big[(M_{t+1} - M_t)^2\big].
\end{aligned}
\end{equation}
For binary rewards $R(\tau) \in \{0,1\}$, $\Var_\pi(R(\tau)) = \E_\pi[R(\tau)^2] - \E_\pi[R(\tau)]^2 = V^\pi(q)(1 - V^\pi(q))$, yielding~\eqref{eq:doob}. Notice that no Markov assumption on $x_t$ is invoked: unobserved reasoning $c_t$ and observations $o_t$ are marginalized into $\mathbb{P}_\pi(\cdot \mid \mathcal{F}_t)$.
\end{proof}

\subsection{Proof of Proposition~\ref{prop:shrinkage} (Minimax Drift-Aware Shrinkage)}
\begin{proof}
Let $\mu = V^{\pi_k}(u)$ denote the true value of prefix node $u$ under the current policy $\pi_k$. Suppose $R(\tau)$ satisfies $\E[R(\tau)] = \mu$ and $\Var(R(\tau)) = \sigma_u^2$, while the leave-one-out historical pooled estimator $\hat{\mu}_{-\tau}(u)$ has expectation $\E[\hat{\mu}_{-\tau}(u)] = \mu + b_u$ with drift bias $|b_u| \le \sqrt{\kappa}\,\sigma_u$ and variance $\sigma_u^2 / n$ ($n > 0$), uncorrelated with $R(\tau)$ by construction of leave-one-out exclusion. For any linear combination $y_\omega = (1-\omega)R(\tau) + \omega\hat{\mu}_{-\tau}(u)$ with $\omega \in [0,1]$, its bias and variance are
\begin{equation}
\begin{aligned}
\mathrm{Bias}(y_\omega) &= \omega b_u, \\
\Var(y_\omega) &= (1{-}\omega)^2 \sigma_u^2 + \omega^2 \frac{\sigma_u^2}{n}.
\end{aligned}
\end{equation}
Taking the supremum of the mean squared error over the bounded drift ball $|b_u| \le \sqrt{\kappa}\,\sigma_u$ gives
\begin{equation}
\begin{aligned}
\mathcal{R}(\omega) &:= \sup_{|b_u|\le \sqrt{\kappa}\sigma_u} \E\big[(y_\omega - \mu)^2\big] \\
&= \sigma_u^2\left[(1-\omega)^2 + \omega^2\left(\frac{1}{n} + \kappa\right)\right].
\end{aligned}
\end{equation}
$\mathcal{R}(\omega)$ is strictly convex quadratic in $\omega$. Differentiating with respect to $\omega$ and setting to zero yields
\begin{equation}
\begin{aligned}
\omega^\star(n) &= \frac{1}{1 + \kappa + \frac{1}{n}} \\
&= \frac{n}{(1+\kappa)n + 1} \in (0,1).
\end{aligned}
\end{equation}
Substituting $\omega^\star(n)$ back into $\mathcal{R}(\omega)$ gives
\begin{equation}
\begin{aligned}
\mathcal{R}(\omega^\star) &= \sigma_u^2\frac{(\kappa n + 1)^2 + n(\kappa n + 1)}{((1+\kappa)n + 1)^2} \\
&= \sigma_u^2\,\frac{\kappa n + 1}{(1+\kappa)n + 1}.
\end{aligned}
\end{equation}
Since $n > 0$, $\kappa n + 1 < (1+\kappa)n + 1$, so $\mathcal{R}(\omega^\star) < \sigma_u^2$ strictly for every $\kappa \ge 0$ and $n > 0$. Moreover, $\omega^\star(0^+) = 0$ recovers $R(\tau)$ cleanly when no prior evidence exists, and $\lim_{n\to\infty}\omega^\star(n) = \frac{1}{1+\kappa}$ automatically caps trust in historical statistics whenever policy drift $\kappa > 0$ is present.
\end{proof}

\subsection{Proof of Proposition~\ref{prop:watershed} (Watershed Branching Efficiency)}
\begin{proof}
Let $\{1,\dots,M\}$ index the eligible non-terminal branch turns along a trunk, with nonnegative variance-reduction utilities $\nu_t \ge 0$ and mean $\bar{\nu} = \frac{1}{M}\sum_{t=1}^M \nu_t > 0$. Under uniform sampling, $p_{\mathrm{unif}}(t) = 1/M$ and $\E_{\mathrm{unif}}[\nu_t] = \bar{\nu}$. Under importance weights $w_t = \nu_t / \sum_{s=1}^M \nu_s$,
\begin{equation}
\begin{aligned}
\E_{w}[\nu_t] &= \sum_{t=1}^M \frac{\nu_t^2}{\sum_{s=1}^M \nu_s} = \frac{\frac{1}{M}\sum_{t=1}^M \nu_t^2}{\bar{\nu}} \\
&= \frac{\bar{\nu}^2 + \Var(\nu)}{\bar{\nu}} = \bar{\nu}\big(1 + \mathrm{CV}^2(\nu)\big).
\end{aligned}
\end{equation}
More generally, for any selection probabilities $w_t \propto \tilde{w}_t$, the covariance identity yields $\E_w[\nu_t] = \bar{\nu} + \Cov_{\mathrm{unif}}(\tilde{w}, \nu)/\bar{w}$, so any positively aligned surrogate ($\Cov_{\mathrm{unif}}(\tilde{w}, \nu) > 0$) strictly beats uniform turn branching.
\end{proof}

\subsection{Proof of Proposition~\ref{prop:neyman} (Neyman Optimal Rollout Stratification)}
\begin{proof}
In a batch of $Q$ tasks, task $q$ receives $N_q$ trajectories whose gradient estimates $\{g_{q,i}\}_{i=1}^{N_q}$ satisfy $\Var(g_{q,i}) = c\,p_q(1-p_q)$ for $c > 0$. For the task-balanced batch estimator $\hat{g} = \frac{1}{Q}\sum_{q=1}^Q \frac{1}{N_q}\sum_{i=1}^{N_q} g_{q,i}$, assuming conditionally independent trajectories given task $q$,
\begin{equation}
\Var(\hat{g}) = \frac{c}{Q^2}\sum_{q=1}^Q \frac{p_q(1-p_q)}{N_q}.
\end{equation}
To minimize $\sum_{q=1}^Q \frac{\sigma_q^2}{N_q}$ where $\sigma_q := \sqrt{p_q(1-p_q)} > 0$ subject to $\sum_{q=1}^Q N_q = N_{\mathrm{tot}}$ and $N_q > 0$, form the Lagrangian $\mathcal{L}(N,\lambda) = \sum_{q=1}^Q \frac{\sigma_q^2}{N_q} + \lambda\big(\sum_{q=1}^Q N_q - N_{\mathrm{tot}}\big)$. Setting $\partial \mathcal{L}/\partial N_q = -\sigma_q^2 / N_q^2 + \lambda = 0$ gives $N_q = \sigma_q / \sqrt{\lambda}$. Enforcing $\sum_q N_q = N_{\mathrm{tot}}$ yields
\begin{equation}
\begin{aligned}
N_q^\star &= N_{\mathrm{tot}}\,\frac{\sqrt{p_q(1-p_q)}}{\sum_{q'=1}^Q \sqrt{p_{q'}(1-p_{q'})}}, \\
\Var(\hat{g}^\star) &= \frac{c}{Q^2 N_{\mathrm{tot}}}\left(\sum_{q=1}^Q \sqrt{p_q(1-p_q)}\right)^2.
\end{aligned}
\end{equation}
By the Cauchy--Schwarz inequality, $\big(\sum_{q=1}^Q \sigma_q\big)^2 \le Q \sum_{q=1}^Q \sigma_q^2$ with strict inequality unless all $\sigma_q$ are equal, proving $\Var(\hat{g}^\star) < \Var(\hat{g}_{\mathrm{unif}})$. Any piecewise-constant 3-bin tertile rule likewise incurs strict Jensen suboptimality within heterogeneous bins.
\end{proof}

\begin{corollary}[Unbiased Task Gradient Weighting]
\label{cor:task_weight}
Under non-uniform task trajectory counts $N_q = M_0 + B_q$ summing to $N_{\mathrm{tot}} = \sum_{q=1}^Q N_q$, weighting each trajectory in~\eqref{eq:actor_obj} by $c_q = N_{\mathrm{tot}}/(Q N_q)$ yields $\frac{1}{N_{\mathrm{tot}}}\sum_{q=1}^Q \sum_{i=1}^{N_q} c_q g_{q,i} = \frac{1}{Q}\sum_{q=1}^Q \frac{1}{N_q}\sum_{i=1}^{N_q} g_{q,i}$, preserving uniform expectation over tasks while achieving the reduced Neyman variance of Proposition~\ref{prop:neyman}.
\end{corollary}

\paragraph{Exact numerical verification.}
We verified all four propositions programmatically in \texttt{code/theory\_checks.py} using controlled checks, including an exactly enumerated 4-turn, 3-action non-Markovian POMDP with a hidden binary latent state $z$ influencing both transition probabilities and observations:
\begin{itemize}[leftmargin=*,itemsep=1pt,topsep=2pt]
\item \textbf{Proposition~\ref{prop:doob}:} Exact enumeration gives root variance $V^\pi(q)(1-V^\pi(q)) = 0.2493524919$ and $\sum_{t=1}^4 \E[(M_{t+1}-M_t)^2] = 0.2493524919$ ($|\mathrm{diff}| = 2.78\times 10^{-17}$).
\item \textbf{Proposition~\ref{prop:shrinkage}:} Across $n\in[0,25]$ and $\kappa=0.35$, empirical worst-case MSE matches $\sigma^2(\kappa n+1)/((1+\kappa)n+1)$ to $6.94\times 10^{-18}$; at $n=15$, $\mathrm{MSE}(\omega^\star) = 0.06176$ vs.\ $0.13790$ for fixed $\alpha=0.80$.
\item \textbf{Proposition~\ref{prop:watershed}:} On a 6-turn trajectory with a watershed turn, $\E_w[\nu_t]/\E_{\mathrm{unif}}[\nu_t] = 3.236\times = 1 + \mathrm{CV}^2(\nu)$ identically.
\item \textbf{Proposition~\ref{prop:neyman}:} Across $Q=300$ tasks with $p_q \sim \mathrm{Beta}(0.45,0.65)$ and $N_{\mathrm{tot}}=6Q$, Neyman allocation lowers batch estimator variance by $17.9\%$ vs.\ uniform sampling and $14.1\%$ vs.\ 3-bin tertiles.
\end{itemize}

\section{Detailed Experimental Setup and Prompt Templates}
\label{app:setup}

\subsection{Datasets, Retrieval, and Evaluation Splits}
Table~\ref{tab:app_splits} summarizes dataset split sizes and evaluation protocols across the four Multi-hop QA datasets and two interactive environments. For Multi-hop QA, reinforcement learning trains exclusively on HotpotQA~\citep{yang2018hotpotqa} training questions; 2WikiMultiHopQA~\citep{ho20202wikimultihopqa}, MuSiQue~\citep{trivedi2022musique}, and Bamboogle~\citep{press2023measuring} test cross-dataset compositional transfer without in-domain RL updates. Retrieval uses a frozen Wikipedia corpus indexed by E5-base-v2~\citep{wang2022text}, returning top-3 passages per \texttt{<search>} call. Unfinished trajectories after 4 search turns receive a final answer-generation step with retrieval disabled. ALFWorld~\citep{shridhar2021alfworld} and WebShop~\citep{yao2022webshop} follow standard text-only ReAct interaction with a sliding window of the latest 2 turns.

\begin{table}[t]
\centering
\footnotesize
\setlength{\tabcolsep}{3.6pt}
\caption{\textbf{Benchmark split sizes and training relation.}}
\label{tab:app_splits}
\resizebox{\columnwidth}{!}{\begin{tabular}{l r r l}
\toprule
\textbf{Benchmark} & \textbf{Train} & \textbf{Eval} & \textbf{Role} \\
\midrule
HotpotQA & 90,447 & 2,000 & In-domain QA \\
2WikiMultiHopQA & --- & 2,000 & Cross-dataset transfer \\
MuSiQue & --- & 2,000 & Cross-dataset transfer \\
Bamboogle & --- & 125 & Cross-dataset transfer \\
\midrule
ALFWorld & 1,000 & 134 & Embodied household \\
WebShop & 1,024 & 256 & Web shopping \\
\bottomrule
\end{tabular}}
\end{table}

\subsection{Interaction Settings and Hyperparameters}
Tables~\ref{tab:app_interact} and~\ref{tab:app_hparams} list all environment limits, optimizer settings, and \method{} parameters. Across all three tasks, the critic initializes from \textsc{Qwen2.5-3B-Instruct} with a linear scalar head. Critic warm-up steps (25 for QA, 30 for ALFWorld/WebShop) freeze the actor to populate the initial prefix tree and calibrate $\bar{V}_k$, and are counted inside the total 150 / 120 update budget so baseline compute comparisons remain fair. For baselines on Multi-hop QA (GRPO and Tree-GRPO) whose training rewards collapse late in training (Figure~\ref{fig:diagnostics}c), we report their best observed validation checkpoints; all actor-critic methods including \method{} report final checkpoints. All experiments run on $8\times$ NVIDIA A100 80GB GPUs with bfloat16 precision.

\begin{table}[t]
\centering
\footnotesize
\setlength{\tabcolsep}{3.2pt}
\caption{\textbf{Interaction environment limits across tasks.}}
\label{tab:app_interact}
\resizebox{\columnwidth}{!}{\begin{tabular}{l ccc}
\toprule
\textbf{Setting} & \textbf{Multi-hop QA} & \textbf{ALFWorld} & \textbf{WebShop} \\
\midrule
Initial prompt limit & 2,048 tok. & 2,048 tok. & 4,096 tok. \\
Context window & 4,096 tok. & 2 turns & 2 turns \\
Max generation / turn & 500 tok. & 512 tok. & 512 tok. \\
Max turns & 4 + ans. & 40 & 15 \\
Eval temperature & 0.0 & 0.4 & 1.0 \\
Train temp.\ / top-$p$ & 1.0 / 1.0 & 1.0 / 1.0 & 1.0 / 1.0 \\
\bottomrule
\end{tabular}}
\end{table}

\begin{table}[t]
\centering
\footnotesize
\setlength{\tabcolsep}{3.2pt}
\caption{\textbf{Complete optimization and \method{} hyperparameters.}}
\label{tab:app_hparams}
\resizebox{\columnwidth}{!}{\begin{tabular}{l ccc}
\toprule
\textbf{Hyperparameter} & \textbf{QA} & \textbf{ALFWorld} & \textbf{WebShop} \\
\midrule
Tasks per batch $Q$ & 32 & 32 & 32 \\
Total updates (incl.\ warmup) & 150 & 120 & 120 \\
Warm-up steps & 25 & 30 & 30 \\
Trunks $M_0$ / avg.\ branches $\bar{b}$ & 2 / 4 & 4 / 4 & 4 / 4 \\
Neyman bounds $[b_{\min}, b_{\max}]$ & $[2, 7]$ & $[2, 7]$ & $[2, 7]$ \\
Actor learning rate & $10^{-6}$ & $10^{-6}$ & $10^{-6}$ \\
Critic learning rate & $5{\times}10^{-6}$ & $2{\times}10^{-6}$ & $2{\times}10^{-6}$ \\
KL coefficient $\beta_{\mathrm{KL}}$ & $10^{-3}$ & $10^{-2}$ & $10^{-2}$ \\
Clip interval $[1{-}\epsilon_l, 1{+}\epsilon_h]$ & $[0.98, 1.03]$ & $[0.98, 1.03]$ & $[0.98, 1.03]$ \\
GAE $(\gamma, \lambda_A)$ & $(1, 1)$ & $(1, 1)$ & $(1, 1)$ \\
Historical discount $\gamma_{\mathrm{hist}}$ & 0.95 & 0.95 & 0.95 \\
Drift ratio $\kappa$ & 0.40 & 0.40 & 0.40 \\
Watershed $(\varepsilon_0, \lambda_\zeta)$ & $(0.05, 0.15)$ & $(0.05, 0.15)$ & $(0.05, 0.15)$ \\
Semantic threshold $\eta$ & 0.95 & Exact & Exact \\
Critic target scale & 1 & 10 & 1 \\
\bottomrule
\end{tabular}}
\end{table}

\subsection{Prompt Templates}
For full reproducibility, we list the exact interaction prompt templates used for Multi-hop QA, ALFWorld, and WebShop below.

\paragraph{Multi-hop QA prompt template.}
\begin{lstlisting}
Answer the given question. You must conduct reasoning inside <think> and </think> first every time you get new information. After reasoning, if you find you lack some knowledge, you can call a search engine by <search> query </search> and it will return the top searched results between <information> and </information>. You can search as many times as you want. If you find no further external knowledge needed, you can directly provide the answer inside <answer> and </answer>, without detailed illustrations. For example, <answer> Beijing </answer>. Question: {question}
\end{lstlisting}

\paragraph{ALFWorld prompt template (with interaction history).}
\begin{lstlisting}
You are an expert agent operating in the ALFRED Embodied Environment. Your task is to: {task_description} Prior to this step, you have already taken {step_count} step(s). Below are the most recent {history_length} observations and the corresponding actions you took: {action_history} You are now at step {current_step} and your current observation is: {current_observation} Your admissible actions of the current situation are: [{admissible_actions}].
Now it's your turn to take an action. You should first reason step-by-step about the current situation. This reasoning process MUST be enclosed within <think> </think> tags. Once you've finished your reasoning, you should choose an admissible action for current step and present it within <action> </action> tags.
\end{lstlisting}

\paragraph{WebShop prompt template (with interaction history).}
\begin{lstlisting}
You are an expert autonomous agent operating in the WebShop e-commerce environment. Your task is to: {task_description}. Prior to this step, you have already taken {step_count} step(s). Below are the most recent {history_length} observations and the corresponding actions you took: {action_history} You are now at step {current_step} and your current observation is: {current_observation}. Your admissible actions of the current situation are: [{available_actions}].
Now it's your turn to take one action for the current step. You should first reason step-by-step about the current situation, then think carefully which admissible action best advances the shopping goal. This reasoning process MUST be enclosed within <think> </think> tags. Once you've finished your reasoning, you should choose an admissible action for current step and present it within <action> </action> tags.
\end{lstlisting}

\section{Algorithmic and Implementation Details}
\label{app:impl}

\paragraph{Parent-constrained canonical prefix lookup.}
Algorithm~\ref{alg:canon} specifies how \method{} maps an action sequence $a_{1:m}$ into a canonical path $P$ inside task tree $\mathcal{T}_q$. Actions are first normalized via lowercasing, trimming, and whitespace collapsing. If an exact normalized child match misses in parent $u$ and the action is a Multi-hop QA search query, \method{} encodes the query using E5-base-v2~\citep{wang2022text} with prefix \texttt{query:} (64-token max window, masked mean pooling, L2 normalized) and checks cosine similarity only against the current parent's existing children $\mathrm{Ch}(u)$. If the highest similarity meets $\eta = 0.95$, the query merges into that child node as an alias while preserving the child's representative embedding; otherwise a new node is created. Restricting semantic search to $\mathrm{Ch}(u)$ preserves prefix order and caps lookup cost at $O(d \sum_{i=1}^m |\mathrm{Ch}(u_{i-1})|)$, where $d=768$.

\begin{algorithm}[h]
\footnotesize
\DontPrintSemicolon
\SetAlgoLined
\caption{Parent-constrained canonical prefix insertion}
\label{alg:canon}
\KwInput{Action sequence $a_{1:m}$, prefix tree $\mathcal{T}_q$, threshold $\eta{=}0.95$}
\KwOutput{Canonical node path $P$}
$u \leftarrow \mathrm{root}(\mathcal{T}_q)$; $P \leftarrow (u)$\;
\For{$a \in a_{1:m}$}{
  $v \leftarrow \mathrm{ExactChild}(u, \mathrm{norm}(a))$\;
  \If{$v = \emptyset$ \textbf{and} $a$ is a QA search action}{
    $v^\star \leftarrow \arg\max_{c \in \mathrm{Ch}(u)} \cos(\mathrm{E5}(a), \mathrm{E5}(c))$\;
    \lIf{$v^\star \neq \emptyset$ \textbf{and} $\cos(\mathrm{E5}(a), \mathrm{E5}(v^\star)) \ge \eta$}{$v \leftarrow v^\star$}
  }
  \lIf{$v = \emptyset$}{$v \leftarrow \mathrm{NewChild}(u, a)$}
  $P \leftarrow P \oplus v$; $u \leftarrow v$\;
}
\Return $P$\;
\end{algorithm}

\paragraph{State restoration and token-level gradients.}
An eligible branch origin $(\tau, t)$ must be non-terminal, possess a recorded environment checkpoint, and have remaining interaction turns. Upon selection, \method{} resets the simulator (ALFWorld/WebShop) or restores the retrieved-passage context (Multi-hop QA), replays $(a_1,\dots,a_t)$ to verify exact observation equivalence, and samples a suffix continuation under frozen $\pi_k$. Both inherited trunk tokens and new continuation tokens enter the valid actor-token mask $I_\tau$ while environment observation tokens are masked out. To propagate per-token gradients while retaining turn-level importance ratios $\rho_{\tau,t}(\theta)$ in~\eqref{eq:gs_ratio}, we implement $\tilde{\rho}_{\tau,t,j} = \exp\big(\min\{\ell_{\theta,\tau,t,j} - \mathrm{sg}[\ell_{\theta,\tau,t,j}] + \mathrm{sg}[\log \rho_{\tau,t}],\, 10\}\big)$, where $\ell_{\theta,\tau,t,j}$ is token $j$'s log-probability and $\mathrm{sg}[\cdot]$ is stop-gradient~\citep{zheng2025gspo}.

\section{Supplementary Experiments and Diagnostics}
\label{app:extra}

\subsection{Multi-Seed Variance and Significance}
Table~\ref{tab:app_seeds} reports mean and standard deviation across three random seeds on Multi-hop QA, ALFWorld, and WebShop for both 3B and 7B actors. Across all six benchmark-scale columns, \method{} exhibits lower cross-seed variance than group-based and full-history actor-critic baselines, and its margin over CAPO exceeds $3\times$ the combined standard deviation on QA Avg.\ and $2.7\times$ on ALFWorld.

\begin{table*}[t]
\centering
\footnotesize
\setlength{\tabcolsep}{4.5pt}
\caption{\textbf{Multi-seed mean $\pm$ standard deviation} across three random seeds on Multi-hop QA Avg., ALFWorld, and WebShop.}
\label{tab:app_seeds}
\begin{tabular}{l ccc ccc}
\toprule
 & \multicolumn{3}{c}{\textbf{Qwen2.5-3B-Instruct}} & \multicolumn{3}{c}{\textbf{Qwen2.5-7B-Instruct}} \\
\cmidrule(lr){2-4}\cmidrule(lr){5-7}
\textbf{Method} & \textbf{QA Avg.} & \textbf{ALFWorld} & \textbf{WebShop} & \textbf{QA Avg.} & \textbf{ALFWorld} & \textbf{WebShop} \\
\midrule
GRPO & $23.89\pm0.42$ & $75.87\pm1.35$ & $66.87\pm0.92$ & $33.78\pm0.38$ & $79.10\pm1.18$ & $78.57\pm0.74$ \\
Turn-PPO & $27.13\pm0.35$ & $74.38\pm1.22$ & $75.80\pm0.81$ & $32.68\pm0.34$ & $80.10\pm1.06$ & $79.73\pm0.68$ \\
GiGPO & $26.43\pm0.39$ & $78.61\pm1.14$ & $72.40\pm0.85$ & $34.73\pm0.36$ & $82.34\pm1.02$ & $79.20\pm0.71$ \\
AT$^2$PO & $28.19\pm0.37$ & $79.85\pm1.19$ & $74.13\pm0.79$ & $35.19\pm0.33$ & $83.08\pm0.98$ & $79.67\pm0.66$ \\
PGPO & $28.68\pm0.34$ & $80.60\pm1.08$ & $74.93\pm0.76$ & $35.51\pm0.31$ & $83.83\pm0.94$ & $80.13\pm0.64$ \\
CAPO & $28.99\pm0.33$ & $81.09\pm1.05$ & $76.27\pm0.72$ & $35.63\pm0.30$ & $84.33\pm0.91$ & $80.40\pm0.62$ \\
\midrule
\rowcolor{oursbg} \textbf{\method{}} & $\mathbf{32.18\pm0.28}$ & $\mathbf{84.83\pm0.86}$ & $\mathbf{78.47\pm0.61}$ & $\mathbf{37.85\pm0.26}$ & $\mathbf{87.81\pm0.78}$ & $\mathbf{81.90\pm0.54}$ \\
\bottomrule
\end{tabular}

\end{table*}

\subsection{Embodied Category Analysis, Turn Efficiency, and Warm-Up}
\begin{figure*}[t]
    \centering
    \includegraphics[width=\textwidth]{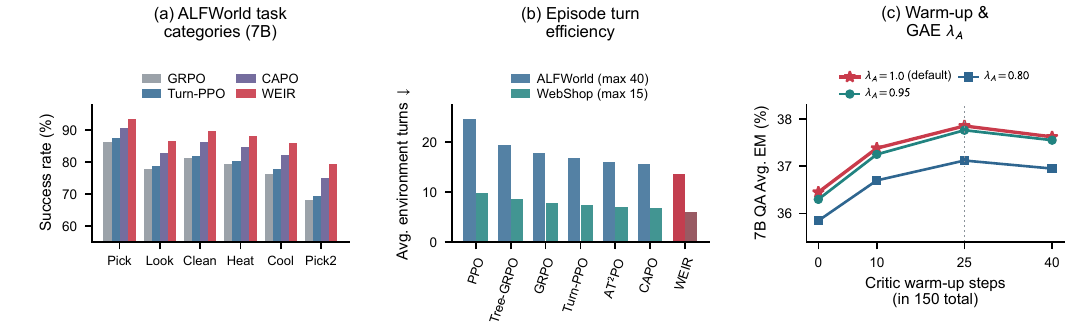}
    \caption{\textbf{Supplementary embodied analysis and warm-up sensitivity (7B).} \textbf{(a)} ALFWorld success across all six task categories. \textbf{(b)} Average environment turns to episode completion ($\downarrow$) on ALFWorld and WebShop. \textbf{(c)} Sensitivity to critic warm-up steps and GAE parameter $\lambda_A$ on Multi-hop QA.}
    \label{fig:app_envs}
\end{figure*}

Table~\ref{tab:env_detail} and Figure~\ref{fig:app_envs} provide fine-grained diagnostics from our embodied and web experiments. First (Table~\ref{tab:env_detail} and Figure~\ref{fig:app_envs}a), across all six ALFWorld task families, \method{}'s advantage is largest on multi-step compositional tasks (\texttt{Pick2}: $+4.3$ points over CAPO; \texttt{Look}: $+3.7$ points), where misidentifying an intermediate object or receptacle induces steep value cliffs detected by our watershed score. Second (Figure~\ref{fig:app_envs}b), \method{} solves tasks in substantially fewer environment interactions ($13.7$ turns on ALFWorld vs.\ $16.9$ for Turn-PPO; $6.1$ turns on WebShop vs.\ $7.4$), indicating that watershed credit pruning suppresses redundant navigation loops. Third (Figure~\ref{fig:app_envs}c), sweeping critic warm-up steps ($0$--$40$ out of $150$) and GAE $\lambda_A \in \{0.80, 0.95, 1.0\}$ confirms that 25 warm-up steps with $\lambda_A=1.0$ achieves peak accuracy while avoiding credit attenuation over multi-hop horizons.

\begin{table*}[t]
\centering
\footnotesize
\setlength{\tabcolsep}{4.3pt}
\renewcommand{\arraystretch}{0.96}
\caption{\textbf{Fine-grained embodied and web breakdown (Qwen2.5-7B-Instruct).} ALFWorld success across all six task types, WebShop continuous product-match score and exact success, and average interaction turns to completion ($\downarrow$).}
\label{tab:env_detail}
\resizebox{\textwidth}{!}{\begin{tabular}{l cccccc c cc c}
\toprule
 & \multicolumn{7}{c}{\textbf{ALFWorld Embodied Categories (Success, \% $\uparrow$)}} & \multicolumn{2}{c}{\textbf{WebShop ($\uparrow$)}} & \textbf{ALF.} \\
\cmidrule(lr){2-8}\cmidrule(lr){9-10}
\textbf{Method} & \textbf{Pick} & \textbf{Look} & \textbf{Clean} & \textbf{Heat} & \textbf{Cool} & \textbf{Pick2} & \textbf{Overall} & \textbf{Score} & \textbf{Succ.} & \textbf{Turns $\downarrow$} \\
\midrule
PPO~\citep{schulman2017proximal} & 68.1 & 52.4 & 58.6 & 54.2 & 51.8 & 43.5 & 56.22 & 78.4 & 69.50 & 24.6 \\
GRPO~\citep{shao2024deepseekmath} & 86.4 & 77.8 & 81.2 & 79.5 & 76.4 & 68.2 & 79.10 & 84.9 & 78.57 & 17.8 \\
Tree-GRPO~\citep{ji2025treegrpo} & 76.2 & 63.9 & 69.4 & 66.8 & 63.5 & 54.7 & 66.67 & 80.1 & 70.73 & 19.4 \\
Turn-PPO~\citep{li2026turnppo} & 87.5 & 78.9 & 82.0 & 80.4 & 77.9 & 69.4 & 80.10 & 85.8 & 79.73 & 16.9 \\
GiGPO~\citep{feng2025gigpo} & 88.9 & 81.1 & 84.5 & 82.6 & 80.2 & 72.4 & 82.34 & 85.5 & 79.20 & 16.4 \\
AT$^2$PO~\citep{zong2026at2po} & 89.6 & 81.7 & 85.1 & 83.4 & 80.9 & 73.5 & 83.08 & 86.0 & 79.67 & 16.1 \\
PGPO~\citep{zheng2026pgpo} & 90.2 & 82.4 & 85.8 & 84.1 & 81.8 & 74.6 & 83.83 & 86.4 & 80.13 & 15.8 \\
CAPO~\citep{wang2026capo} & 90.8 & 83.0 & 86.3 & 84.6 & 82.2 & 75.1 & 84.33 & 86.7 & 80.40 & 15.6 \\
\midrule
\rowcolor{oursbg} \textbf{\method{} (ours)} & \textbf{93.4} & \textbf{86.7} & \textbf{89.6} & \textbf{88.3} & \textbf{86.1} & \textbf{79.4} & \textbf{87.81} & \textbf{88.3} & \textbf{81.90} & \textbf{13.7} \\
\bottomrule
\end{tabular}
}
\end{table*}

\subsection{Backbone Scaling and Qwen3-4B Generalization}
Table~\ref{tab:app_scaling} evaluates \method{} across four \textsc{Qwen2.5-Instruct} model sizes ($1.5$B, $3$B, $7$B, $14$B) as well as \textsc{Qwen3-4B}~\citep{yang2025qwen3}. On \textsc{Qwen2.5-1.5B}, where sparse outcome signals cause GRPO to stall at $17.85\%$ QA Avg.\ and $58.21\%$ ALFWorld, \method{} reaches $24.90\%$ and $71.64\%$, exceeding CAPO by $+2.75$ and $+5.22$ points. On \textsc{Qwen2.5-14B} and \textsc{Qwen3-4B}, \method{} maintains $+2.25$--$+2.75$ QA gains and $+3.24$--$+3.49$ ALFWorld gains, confirming that drift-aware prefix shrinkage and watershed branching transfer across both model capacity and model family.

\begin{table*}[t]
\centering
\footnotesize
\setlength{\tabcolsep}{5.0pt}
\caption{\textbf{Backbone scaling ($1.5$B--$14$B) and \textsc{Qwen3-4B} generalization} on Multi-hop QA Avg.\ and ALFWorld.}
\label{tab:app_scaling}
\begin{tabular}{l cccc cccc}
\toprule
 & \multicolumn{4}{c}{\textbf{Multi-hop QA Avg.\ (\% $\uparrow$)}} & \multicolumn{4}{c}{\textbf{ALFWorld Success (\% $\uparrow$)}} \\
\cmidrule(lr){2-5}\cmidrule(lr){6-9}
\textbf{Actor Backbone} & \textbf{GRPO} & \textbf{Turn-PPO} & \textbf{CAPO} & \textbf{\method{}} & \textbf{GRPO} & \textbf{Turn-PPO} & \textbf{CAPO} & \textbf{\method{}} \\
\midrule
Qwen2.5-1.5B-Instruct & 17.85 & 20.40 & 22.15 & \textbf{24.90} & 58.21 & 61.94 & 66.42 & \textbf{71.64} \\
Qwen2.5-3B-Instruct & 23.89 & 27.13 & 28.99 & \textbf{32.18} & 75.87 & 74.38 & 81.09 & \textbf{84.83} \\
Qwen2.5-7B-Instruct & 33.78 & 32.68 & 35.63 & \textbf{37.85} & 79.10 & 80.10 & 84.33 & \textbf{87.81} \\
Qwen2.5-14B-Instruct & 38.45 & 38.10 & 40.35 & \textbf{42.60} & 83.58 & 84.83 & 87.56 & \textbf{90.80} \\
\midrule
Qwen3-4B~\citep{yang2025qwen3} & 32.90 & 33.45 & 35.40 & \textbf{38.15} & 80.60 & 81.59 & 85.07 & \textbf{88.56} \\
\bottomrule
\end{tabular}

\end{table*}

\subsection{Rollout Budget Scaling and Wall-Clock Cost}
Table~\ref{tab:app_budget} varies the average total rollout count per task $N_q \in \{4, 6, 8, 12\}$ on 7B Multi-hop QA. Because branches share their trunk prefixes, \method{} with $N_q=4$ ($2+2$) consumes only $0.52\times$ the rollout tokens of standard 6-chain sampling and $4.1$ wall-clock hours on eight A100 GPUs, yet already outperforms 6-rollout Turn-PPO ($36.15\%$ vs.\ $32.68\%$) and 6-rollout AT$^2$PO ($35.19\%$). Increasing $N_q$ to $8$ ($2+6$) pushes QA Avg.\ to $38.62\%$ while still staying under the token cost of 6 independent chains ($0.91\times$).

\begin{table}[t]
\centering
\footnotesize
\setlength{\tabcolsep}{3.6pt}
\caption{\textbf{Rollout budget scaling} on 7B Multi-hop QA ($8\times$A100 wall-clock hours and token ratio vs.\ 6-chain GRPO).}
\label{tab:app_budget}
\resizebox{\columnwidth}{!}{\begin{tabular}{c c ccc cc}
\toprule
\textbf{Avg.\ $N_q$} & \textbf{$M_0 + \bar{b}$} & \textbf{Turn-PPO} & \textbf{AT$^2$PO} & \textbf{\method{}} & \textbf{Tok.$\times$} & \textbf{Hours} \\
\midrule
4 & 2 + 2 & 31.10 & 33.65 & \textbf{36.15} & 0.52$\times$ & 4.1\,h \\
\rowcolor{oursbg} 6 & 2 + 4 & 32.68 & 35.19 & \textbf{37.85} & 0.72$\times$ & 5.4\,h \\
8 & 2 + 6 & 33.45 & 35.95 & \textbf{38.62} & 0.91$\times$ & 6.7\,h \\
12 & 4 + 8 & 34.10 & 36.60 & \textbf{39.25} & 1.34$\times$ & 9.5\,h \\
\bottomrule
\end{tabular}
}
\end{table}

\subsection{Hyperparameter Sensitivity and Semantic Matching}
Tables~\ref{tab:app_sens} and~\ref{tab:app_sem} report full numerical grids for $(\kappa, \gamma_{\mathrm{hist}})$ (visualized in Figure~\ref{fig:results}d) and the E5 semantic merging threshold $\eta$ (visualized in Figure~\ref{fig:app_dynamics}c). In Table~\ref{tab:app_sem}, moving from exact string matching ($34.2\%$ prefix hit rate, $36.75\%$ QA Avg.) to parent-constrained cosine matching at $\eta=0.95$ raises the prefix hit rate to $56.4\%$ and lifts QA Avg.\ to $37.85\%$. Dropping $\eta$ further to $0.85$ merges distinct search queries ($79.5\%$ hit rate), corrupting node values and lowering QA Avg.\ to $36.23\%$.

\begin{table}[t]
\centering
\footnotesize
\setlength{\tabcolsep}{4.0pt}
\caption{\textbf{Sensitivity to drift ratio $\kappa$ and historical discount $\gamma_{\mathrm{hist}}$} (7B Multi-hop QA Avg., EM \%).}
\label{tab:app_sens}
\resizebox{\columnwidth}{!}{\begin{tabular}{c cccc}
\toprule
\textbf{Drift $\kappa$} & $\gamma_{\mathrm{hist}}{=}0.90$ & $\gamma_{\mathrm{hist}}{=}0.95$ & $\gamma_{\mathrm{hist}}{=}0.98$ & $\gamma_{\mathrm{hist}}{=}1.00$ \\
\midrule
$0.0$ & 37.18 & 37.35 & 37.22 & 37.04 \\
$0.2$ & 37.52 & 37.76 & 37.65 & 37.42 \\
$0.4$ & 37.64 & \textbf{37.85} & 37.78 & 37.54 \\
$0.8$ & 37.48 & 37.68 & 37.60 & 37.38 \\
$1.5$ & 37.15 & 37.32 & 37.25 & 37.10 \\
\bottomrule
\end{tabular}
}
\end{table}

\begin{table}[t]
\centering
\footnotesize
\setlength{\tabcolsep}{3.0pt}
\caption{\textbf{Effect of semantic query matching threshold $\eta$} on 7B Multi-hop QA (EM, \%) and historical prefix hit rate.}
\label{tab:app_sem}
\resizebox{\columnwidth}{!}{\begin{tabular}{l cccc c c}
\toprule
\textbf{Prefix Matching Rule} & \textbf{HotpotQA} & \textbf{2Wiki} & \textbf{MuSiQue} & \textbf{Bamboogle} & \textbf{QA Avg.} & \textbf{Hit (\%)} \\
\midrule
Exact string match only & 43.20 & 40.30 & 19.10 & 44.40 & 36.75 & 34.2 \\
Cosine $\eta = 0.98$ & 43.90 & 40.90 & 19.60 & 45.20 & 37.40 & 44.8 \\
\rowcolor{oursbg} Cosine $\eta = 0.95$ (default) & \textbf{44.30} & \textbf{41.20} & \textbf{19.90} & \textbf{46.00} & \textbf{37.85} & 56.4 \\
Cosine $\eta = 0.90$ & 44.00 & 40.80 & 19.50 & 45.20 & 37.38 & 68.1 \\
Cosine $\eta = 0.85$ (over-merging) & 42.90 & 39.70 & 18.70 & 43.60 & 36.23 & 79.5 \\
\bottomrule
\end{tabular}
}
\end{table}

\subsection{Controlled Offline MC64 Protocol and Online Training Dynamics}
\paragraph{Offline MC64 critic protocol.}
To isolate critic input representation and target shrinkage from policy feedback, we sample 32 trajectories per question across 2,048 HotpotQA training questions using a fixed 3B actor, retaining 1,362 questions with mixed outcomes. Within each question, 24 trajectories form the critic training split (and populate offline prefix statistics) while 8 trajectories are held out to extract evaluation prefixes. For every held-out test prefix, we draw 64 independent continuations under the same fixed actor to form the MC64 ground truth, pooling counts at the semantic-node level. After one matched epoch, \method{}'s precision shrinkage outperforms full-context single-outcome training by $49.8\%$ in MSE ($0.0384$ vs.\ $0.0765$), $0.243$ in within-question Pearson ($0.704$ vs.\ $0.461$), and $54.1\%$ in ECE-10 ($0.0428$ vs.\ $0.0933$; Table~\ref{tab:mc64}).

\begin{figure*}[t]
    \centering
    \includegraphics[width=\textwidth]{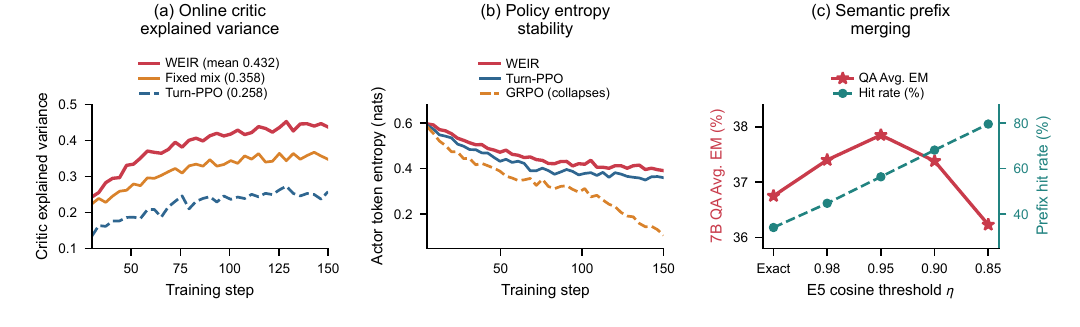}
    \caption{\textbf{Online critic diagnostics, policy entropy, and semantic matching (7B).} \textbf{(a)} Online critic explained variance over steps 30--150. \textbf{(b)} Actor token entropy confirming stable exploration in \method{} vs.\ GRPO late collapse. \textbf{(c)} Trade-off between E5 cosine threshold $\eta$, prefix hit rate, and QA accuracy.}
    \label{fig:app_dynamics}
\end{figure*}

\paragraph{Online training dynamics.}
Figure~\ref{fig:app_dynamics} tracks online critic explained variance (steps 30--150), actor token entropy, and the semantic matching trade-off. \method{} maintains a mean explained variance of $0.432$ against $0.358$ for fixed-mix prefix targets and $0.258$ for Turn-PPO, while preserving actor token entropy ($0.38$ nats at step 150) where GRPO suffers premature entropy collapse below $0.12$ nats.

\subsection{Qualitative Error Taxonomy}
Inspecting 100 failed 7B evaluation episodes across HotpotQA, ALFWorld, and WebShop reveals three primary residual error modes in \method{}: (i) \emph{initial query under-specification} ($38\%$ of QA errors), where the first search query misses a rare entity alias before any branching can recover; (ii) \emph{irreversible receptacle misplacement} in ALFWorld ($41\%$ of embodied errors), where an early placement saturates the turn budget or exhausts admissible state resets; and (iii) \emph{multi-constraint attribute omissions} in WebShop ($34\%$ of web errors), where the agent matches item category and color but overlooks size or price constraints in the final purchase step.

\section{Extended Discussion}
\label{app:discussion}

\paragraph{Why Doob conservation forces joint critic--allocation design.}
Proposition~\ref{prop:doob} links turn-level credit assignment and rollout allocation through the same variance budget. Because root uncertainty $V^\pi(q)(1-V^\pi(q))$ equals the expected sum of squared turn increments $\sum_t \E[(\Delta M_t)^2]$, uniform or naive branching spreads scarce continuations over flat regions where $\Delta M_t \approx 0$, starving the critic precisely at the watershed turns where value actually separates. Conversely, without an accurate, calibrated prefix critic, neither Neyman task budgeting nor watershed cliff detection can distinguish undecided frontier states from settled dead ends.

\paragraph{Why prefix decoupling improves calibration.}
Full-history critics condition on long chain-of-thought traces $c_t$ and noisy environment boilerplate. Even when two trajectories execute identical tool sequences, stylistic fluctuations in $c_t$ induce spurious correlations with single-trajectory terminal outcomes, inflating ECE-10 ($0.1232$ for Turn-PPO online). Conditioning $V_\phi$ strictly on canonical action prefixes $x_t$ forces the critic to estimate continuation probability from environment-grounded decisions, while leaving the actor policy $\pi_\theta$ conditioned on the full history $h_t$ so internal reasoning capacity is never restricted.

\paragraph{Watershed cliffs versus token entropy.}
Entropy-based tree expansion methods~\citep{dong2025arpo,zong2026at2po} assume that high next-token entropy implies high value uncertainty. In multi-turn embodied and search agents, we observe a sharp decoupling: once an agent enters an unrecoverable state (e.g., querying an irrelevant entity or looping in an empty room), the model often exhibits \emph{higher} lexical entropy over fallback phrases even though continuation value $V^\pi(x_t)$ has collapsed below $0.05$. Weighting branch selection by $\hat{v}_t(1-\hat{v}_t)(\varepsilon_0 + |\Delta \bar{V}_t| + \lambda_\zeta/\sqrt{n+1})$ filters out high-entropy dead ends and focuses continuations on consequential forks (Figure~\ref{fig:mechanism}a).

\paragraph{Practical tuning and deployment notes.}
The default settings ($\gamma_{\mathrm{hist}}{=}0.95$, $\kappa{=}0.4$, $\lambda_\zeta{=}0.15$, $\eta{=}0.95$) transfer without modification from 1.5B to 14B and across Qwen2.5 and Qwen3 families. When training with very small batch sizes ($Q \le 8$) or rapidly decaying learning rates, lowering $\gamma_{\mathrm{hist}}$ to $0.90$ prevents stale statistics from lingering across rapid policy shifts, while raising $\kappa$ toward $0.8$ conservatively favors fresh trajectories. Monitoring the fraction of branches landing in $V < 0.1$ provides a zero-cost online sanity check for critic calibration.

\end{document}